\documentclass[preprint,12pt]{elsarticle}

\usepackage{graphicx}

\usepackage{subfig}
\usepackage{booktabs}
\usepackage{multirow}

\usepackage[linesnumbered,ruled,vlined,onelanguage]{algorithm2e}

\usepackage{amsmath,amssymb,amsfonts}
\usepackage{amsthm}
\usepackage{setspace}
\usepackage{float} 

\usepackage{tabularx}
\usepackage{hyperref}
\usepackage[ruled,linesnumbered]{algorithm2e}
\usepackage{url}
\DeclareUnicodeCharacter{0306}{\v{Z}}
\usepackage{enumitem}
\usepackage{textcomp}
\usepackage{xcolor}
\def\BibTeX{{\rm B\kern-.05em{\sc i\kern-.025em b}\kern-.08em
    T\kern-.1667em\lower.7ex\hbox{E}\kern-.125emX}}

\journal{Computer Security}

\begin{document}

\begin{frontmatter}



\title{CRSFL: Cluster-based Resource-aware Split Federated Learning for Continuous Authentication}

\author[ets,lau]{Mohamad Wazzeh}
\author[ets,lau]{Mohamad Arafeh}
\author[ets,lau]{Hani Sami}
\author[uqtr]{Hakima Ould-Slimane}
\author[ets]{Chamseddine Talhi}
\author[khalifa6g,lau]{Azzam Mourad}
\author[khalifacyber]{Hadi Otrok}

\affiliation[ets]{
    organization={Department of Software and IT engineering, École de Technologie Supérieure (ÉTS)},
    city={Montreal},
    postcode={H3C 1K3},
    state={Quebec},
    country={Canada}
}
\affiliation[lau]{
    organization={Artificial Intelligence and Cyber Systems Research Center, Department of CSM, Lebanese American University},
    city={Beirut},
    country={Lebanon}
}
\affiliation[uqtr]{
    organization={Department of mathematics and computer science, Universite de Quebec a Trois-Rivieres (UQTR)},
    city={Trois-Rivieres},
    postcode={G8Z 4M3},
    state={Quebec},
    country={Canada}
}
\affiliation[khalifa6g]{
    organization={KU 6G Research Center, Department of Computer Science, Khalifa University},
    city={Abu Dhabi},
    country={UAE}
}
\affiliation[khalifacyber]{
    organization={Center of Cyber-Physical Systems (C2PS), Department of Computer Science, Khalifa University},
    city={Abu Dhabi},
    country={UAE}
}

\begin{abstract}
In the ever-changing world of technology, continuous authentication and comprehensive access management are essential during user interactions with a device. Split Learning (SL) and Federated Learning (FL) have recently emerged as promising technologies for training a decentralized Machine Learning (ML) model. With the increasing use of smartphones and Internet of Things (IoT) devices, these distributed technologies enable users with limited resources to complete neural network model training with server assistance and collaboratively combine knowledge between different nodes. In this study, we propose combining these technologies to address the continuous authentication challenge while protecting user privacy and limiting device resource usage. However, the model's training is slowed due to SL sequential training and resource differences between IoT devices with different specifications. Therefore, we use a cluster-based approach to group devices with similar capabilities to mitigate the impact of slow devices while filtering out the devices incapable of training the model. In addition, we address the efficiency and robustness of training ML models by using SL and FL techniques to train the clients simultaneously while analyzing the overhead burden of the process. Following clustering, we select the best set of clients to participate in training through a Genetic Algorithm (GA) optimized on a carefully designed list of objectives. The performance of our proposed framework is compared to baseline methods, and the advantages are demonstrated using a real-life UMDAA-02-FD face detection dataset. The results show that CRSFL, our proposed approach, maintains high accuracy and reduces the overhead burden in continuous authentication scenarios while preserving user privacy.
\end{abstract}




\end{frontmatter}


\section{Introduction}

\begin{figure}[ht]
\centering
{\includegraphics[width=.80\textwidth]{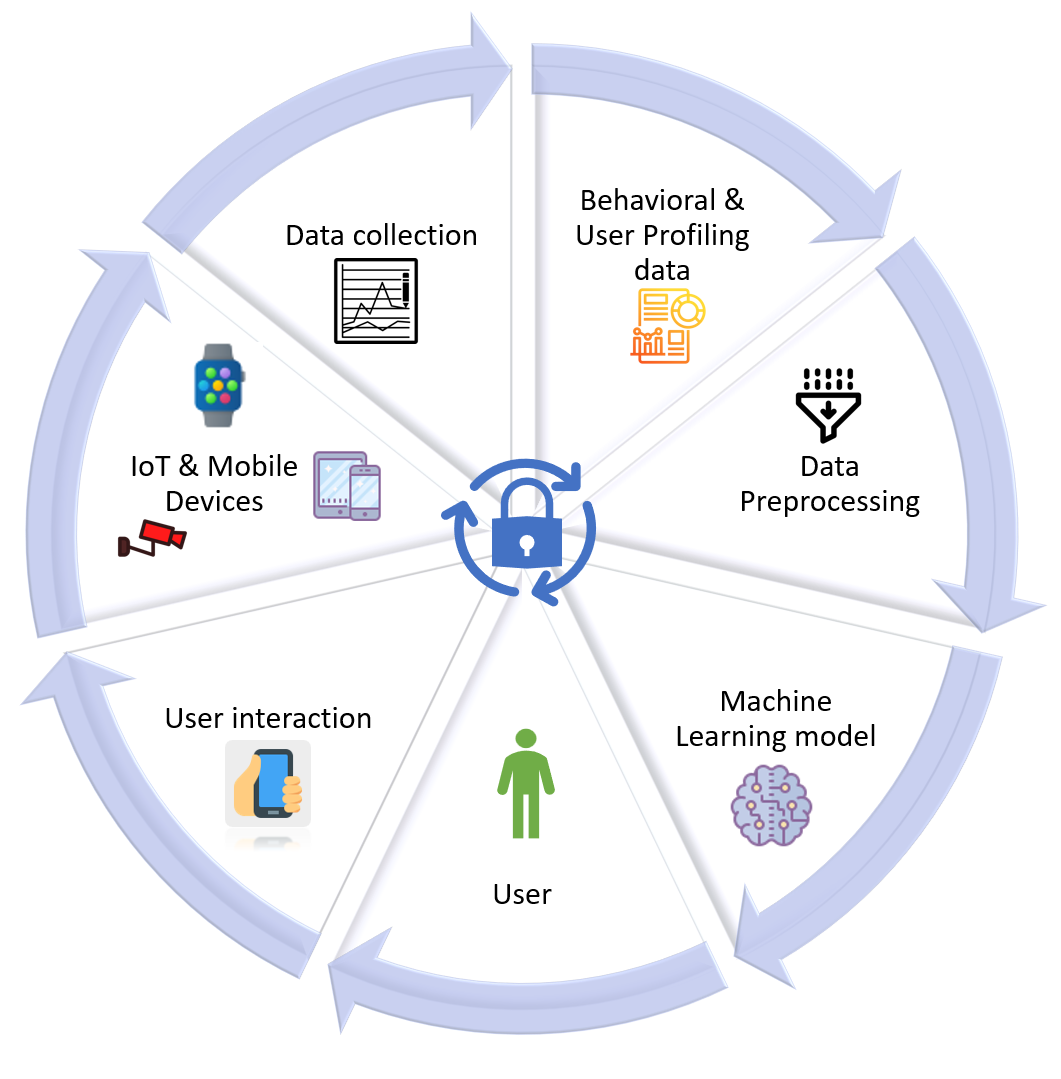}}
\caption{User Interaction and Data Flow for Continuous Authentication in Mobile and IoT Devices.}
\label{fig:iot_mobile_ca}
\end{figure}

Traditional authentication methods face privacy and accuracy challenges when devices are used for extended periods. One-time passwords and PINs are vulnerable to cyber-attacks, such as credential theft, shoulder surfing, and session hijacking \cite{liang2020behavioral}. Continuous authentication is preferred for IoT and mobile devices, which utilize unique physical and behavioural traits like gait analysis, voice, facial images or motion patterns \cite{pritee2024machine}. This method implicitly verifies the user's identity while interacting with IoT devices. Continuous authentication, as shown in Figure \ref{fig:iot_mobile_ca}, involves constantly verifying a person's identity using their unique physical or behavioural characteristics \cite{sanchez2021authcode}. For example, how a person walks or moves while using a device can be used to confirm his/her identity in the background. Users' data is connected to a unique label representing their true identity. This makes it challenging to classify user data accurately using traditional ML tools. The data collected is not independent and identically distributed (non-IID) since data sample numbers and labels differ from one user to another. The classical continuous authentication approaches presented in various studies \cite{lee2017implicit, 4_zhu2019riskcog, lu2018deepauth} transfer behavioural data to an external server to train an ML model. However, such approaches may compromise users' privacy and security, exposing them to cyberattacks like spoofing and data hijacking.
Continuous user authentication is constantly evolving, with new techniques being developed to enhance traditional systems. Numerous studies have utilized distributed learning for continuous user authentication where a multi-class classification model is typically employed to train the users' models \cite{feng2024privacy, wazzeh2022privacy, monschein2021towards, wazzeh2023towards, yazdinejad2021federated, wazzeh2022warmup}. One such approach is federated learning, where clients train local models with their data and transfer weights rather than their raw data to the server for aggregation \cite{mcmahan2017communication}. This approach demonstrates an advantage, as the client's models can be merged on the server end, enabling a singular model to distinguish between various client labels \cite{wahab2021federated, abdulrahman2020survey, truong2021privacy}. In addition, examples of binary federated learning are being utilized for continuous authentication \cite{hosseini2021federated}, wherein a softmax function with the client's model indicates similarity to feature data. However, achieving high accuracy in such cases requires many communication rounds and may not perform well in a distributed learning architecture. 

\begin{figure}[ht]
\centering
{\includegraphics[width=.80\textwidth]{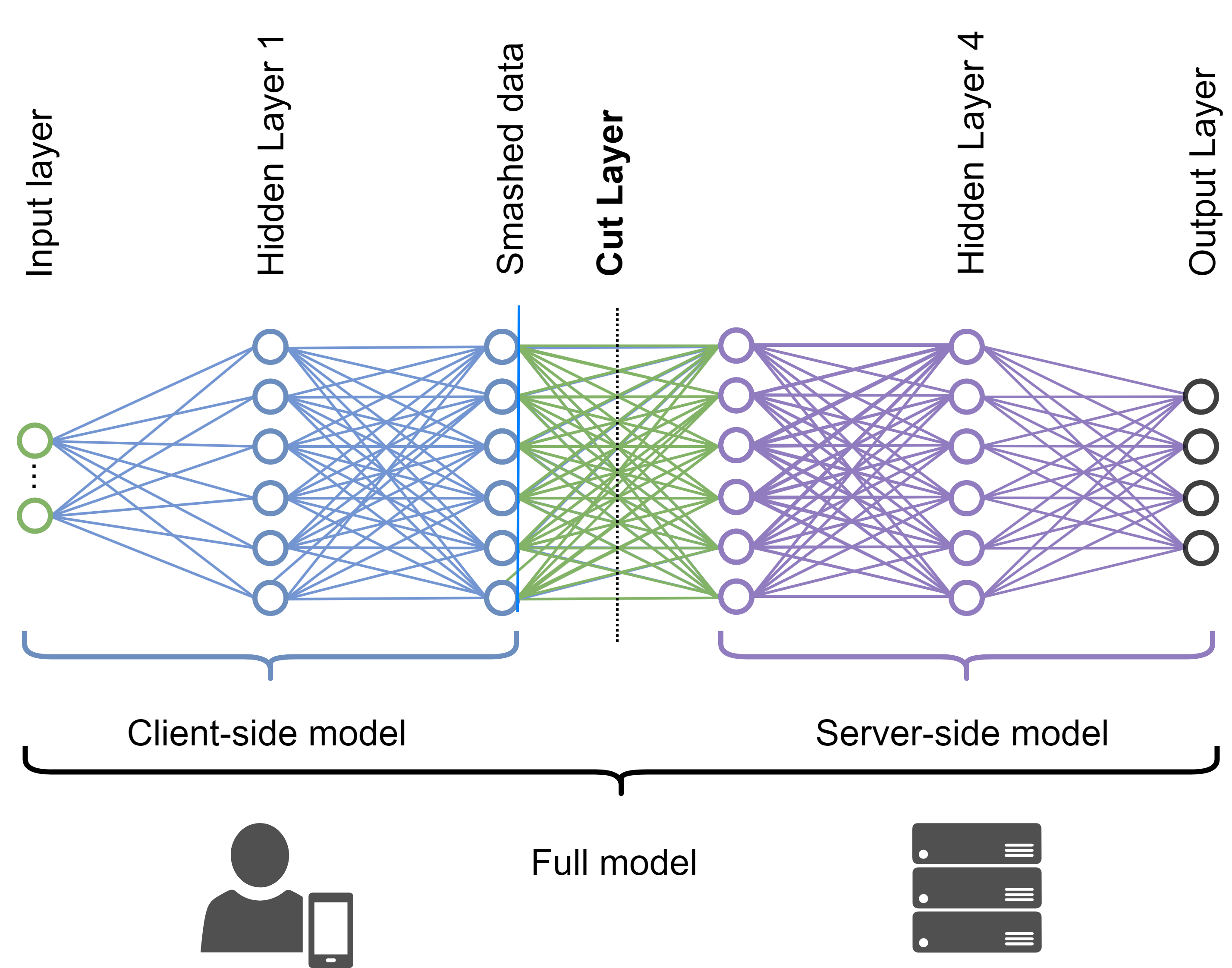}}
\caption{Neural Network Model Splitting.}
\label{fig:sl_model}
\end{figure}

\begin{figure*}[ht]
\centering
{\includegraphics[width=0.99\textwidth]{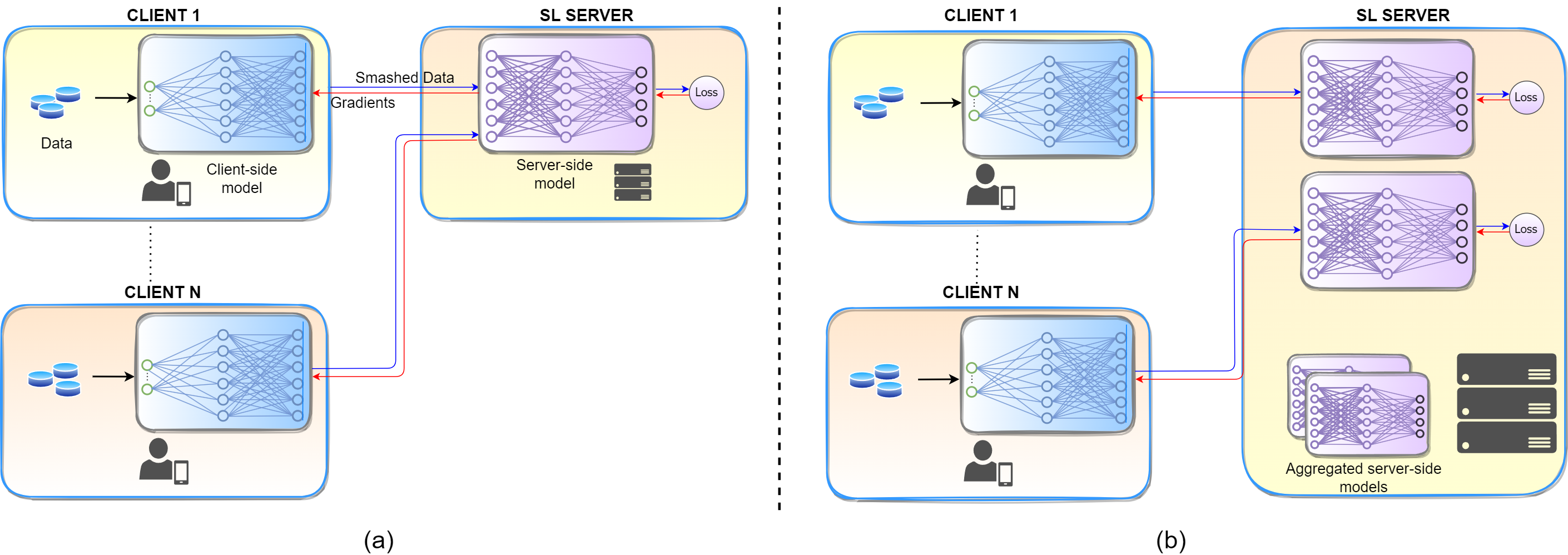}}\hfill 
\caption{A comparison of client-server split learning techniques: (a) Sequential split learning, where each client sends model updates to the server one at a time, and (b) Parallel split learning, where multiple clients send model updates simultaneously to the server.}
\label{fig:sl_sequence_parallel}
\end{figure*}

The FL approach facilitates collaborative training across nodes, sharing knowledge between clients in separate clusters and enhancing the model's resilience. This approach can improve model security and reduce the computational burden on the clients associated with traditional FL methods \cite{gao2021evaluation}. While continuous authentication techniques are typically used in FL \cite{monschein2021towards}, there are some limitations to the existing approaches for FL. Training on IoT devices can sometimes be impractical due to limited resources. Another challenge is that vanilla FL architecture can struggle with model convergence when dealing with unique label characteristics for users \cite{wazzeh2022privacy}. This is because the client models only have access to one label, which can cause weight divergence.

Conversely, SL provides a solution by allowing clients to train only specific neural network layers rather than the entire model, leading to faster training times and reducing the resource burden on the client \cite{vepakomma2018split}. As represented in Figure \ref{fig:sl_model}, SL involves partitioning the neural network model at the cut layer into two parts, with the server undertaking the training process by solely transferring the compressed data of the cut layer weights from the client end, keeping all personal data at the client's device. Additionally, SL's non-aggregation feature has been shown to improve model performance and minimize weight divergence issues between clients \cite{gao2020end}. Nevertheless, employing SL to resolve the problem of continuous authentication could be restricted by concerns related to security and training. This is because SL utilizes a sequential training methodology, which necessitates that each client wait for the training of the other client to terminate and receive model updates before continuing with their training process, as shown in Figure \ref{fig:sl_sequence_parallel}(a).

To overcome the issues of FL and SL and benefit from the strength of each one, we propose in this paper a novel framework for continuous authentication that combines FL and SL and performs multi-class classification to identify users. In contrast to the classical SL method shown in Figure \ref{fig:sl_sequence_parallel}(a), the clients in Figure \ref{fig:sl_sequence_parallel}(b) can train their models with a server in parallel to improve training efficiency \cite{gupta2018distributed, duan2022combined}. Incorporating the strengths of FL and SL can enhance model security, data privacy, and neural network training accuracy \cite{thapa2022splitfed}. The collaborative training process entails a joint effort between the split server, the federated server, and the clients. Each client can train their models in parallel with the split server, utilizing separate server-side weights later aggregated on both servers upon completing the participants' training process round. 

To tackle the challenge of device heterogeneity during training, we have introduced a strategy of organizing devices into clusters. We aim to mitigate the impact of varying computing capabilities across devices by utilizing a clustering technique. This involves grouping devices with similar capabilities, reducing communication overhead and enabling effective collaboration among clients within a cluster during the training process \cite{tu2021feddl, wazzeh2023towards}.

Moreover, when applying clustering, another challenge arises in selecting proper clients to participate in learning. The existing research on continuous authentication using distributed learning lacks advanced client selection techniques and has not thoroughly studied their effects. This can lead to inefficient resource utilization and the missing participation of unique users from the learning model, as the current approaches employ random client selection based on predetermined percentages that are unrealistic when considering the heterogeneity of clients' resource capabilities and the need for a continuous authentication model. Additionally, scalability issues with random and deterministic selection algorithms require further attention. It is crucial to address these challenges to ensure fair resource distribution and optimal performance in these architectures. 
To enhance the adaptability and robustness of resource allocation strategies, we propose, as part of our framework, a heuristic-based algorithm that selects clients based on their changing resource constraints and follows security objectives where each client holds a unique label and has to optimize its participation while maximizing the model performance.
This study introduces a novel approach to achieving continuous user authentication on mobile and IoT devices with improved security and efficiency. In this paper, we describe the structure and algorithms of our proposed method and compare it with existing authentication methods to showcase its effectiveness. Our experimentation demonstrates that our method offers secure and efficient access control for IoT and mobile devices. Our contributions can be summarized as follows:

\begin{itemize}
\item Proposing a Cluster-Based Resource-aware Split Federated Learning (CRSFL) framework tailored for IoT devices. CRSFL ensures users' privacy by maintaining data on their respective devices.

\item Introducing a filtering methodology that utilizes capacity constraints and a clustering process to guarantee clients' engagement with sufficient resource capacities. The constraints are established according to the initial model and device specifications to cluster comparable IoT devices. 

\item Applying a machine learning approach to predict the client device usage of resources. The machine learning evaluates each client's dynamic resource availability in every round and estimates its model training usage. 

\item Employing a heuristic client selection algorithm that optimizes resource allocation through a five-objective scenario based on a Genetic Algorithm (GA). This process considers the client's unique labels, number of devices, event rate, and history of client selection and minimizes wait time during model training rounds for efficient client selection.

\item Conducting extensive experiments on real-life authentication dataset to compare and evaluate our approach to existing methods for continuous authentication. 

\end{itemize}

The experimental results obtained by employing CRSFL reveal its superiority and offer insights for future research in this domain.

\section{Related Work}

Using split and federated learning techniques in decentralized machine learning tasks has shown promising results in improving user authentication on mobile and IoT devices. These methods ensure efficient access control, enhancing the devices'  security and privacy.

\subsection{Federated learning in Continuous authentication}

The authors in \cite{wazzeh2022privacy} proposed a novel technique that uses warm-up models to enhance the performance of authentication models and user verification in federated learning. The authors demonstrated that their approach effectively addresses unique non-IID data and improves model weights. Additionally, in \cite{wazzeh2022warmup}, transfer learning was shown to increase the accuracy and robustness of the models when used in a federated learning approach for continuous authentication. 

The work in \cite{monschein2021towards} explores how to authenticate users within web-based systems by analyzing their behaviour and using machine learning models to detect anomalies. These models are trained federated by multiple organizations, which work together in a peer-to-peer fashion while safeguarding their local data. A data governance component is proposed to ensure that the models and data are high quality and compliant while aligning with the objectives and requirements of all participants involved.

The research work in \cite{yazdinejad2021federated} outlines a drone authentication architecture that utilizes drones' radio frequency (RF) capabilities within IoT networks, leveraging a federated learning approach. The proposed model employs homomorphic encryption and secure aggregation to protect model parameters during data transmission and aggregation. 

The authors in \cite{feng2024privacy} present a privacy-preserving aggregation scheme designed for federated learning in vehicular ad-hoc networks (VANETs). The approach enables collaborative training of a global model without revealing participating vehicles' local data or gradients. The scheme introduces a continuous authentication mechanism based on a non-interactive zero-knowledge proof protocol, which verifies the legitimacy and integrity of clients and their updates. Additionally, the scheme utilizes edge devices to assist with the aggregation process, reducing communication and computation overhead.

The distinction between traditional FL methods and FL tailored for continuous authentication primarily arises from the nature of client data in continuous authentication scenarios. Each client has a unique label in these contexts, with no overlap between clients, causing convergence issues for the model. The works above demonstrate promising results in using a multi-class model in federated learning for continuous authentication. Additionally, while some studies have explored the use of binary models \cite{hosseini2021federated}, it is worth noting that such approaches often require many rounds to achieve high-accuracy results. Nonetheless, it is essential to acknowledge that the capacity of IoT devices to train machine learning models may be restricted by their available resources. Furthermore, the issue of client selection in federated learning for continuous authentication has not been fully explored in the works above. The authors have adopted an approach to client selection based on random or percentage-based criteria without considering available resources in their methodology.

\subsection{Split learning and Federated learning}
The authors in \cite{thapa2022splitfed} employ a split federated learning technique to harness the benefits of distributed computing and parallel client-side training. Their approach ensures that client can train their model in parallel with the server by aggregating the weights.

In the work of \cite{arafeh2023efficient}, the authors proposed a unique double clustering approach to address the challenges posed by non-IID clients and stragglers in the IoT environment. Their approach involves two clustering layers - the first layer focuses on biased clients while the second layer handles stragglers. This method has yielded faster execution times through parallel execution and increased accuracy by clustering non-IID clients into IID clusters. 

In the work in \cite{samikwa2022ares}, a split learning framework is explored to address the issue of slow devices, also known as "stragglers," to minimize training time and energy consumption in IoT devices. The authors implemented a model split point that is adjustable based on the available resources of each device. Their research has demonstrated that this approach effectively reduces energy consumption and mitigates the impact of stragglers.

In \cite{wu2023split}, the authors employed a cluster split federated learning technique to address the issue of training latency for client models in a heterogeneous IoT environment. The authors used a cut layer selection method to create user clusters and optimized client allocation into specific clusters using radio spectrum allocation and channel conditions. 

The authors present a split federated learning approach version in a work of \cite{han2021accelerating}. The authors tackle the challenges of latency and communication efficiency by implementing local loss-based training for split learning, as opposed to the global loss function of the entire model. This approach resulted in comparable accuracy results with a reduced latency.

The above-related work enhancements improve the distributed learning system by incorporating a split learning component, effectively reducing the client's overhead burden. While the authors addressed the communication and channel conditions challenges that come with machine learning model training, they neglected to explore using an advanced selection algorithm for continuous authentication when determining client selection criteria. Our work goes beyond addressing straggling clients and execution time. We have developed a client selection algorithm that optimizes the selection of clients based on their resource capacities and data sample properties using a multi-objective algorithm.

\subsection{Split learning and Federated learning in Continuous authentication}

In their research, \cite{oza2021federated} propose a method for user authentication that utilizes federated and split learning principles. The authors divide the training process between the client and server, but instead of sharing the model weights from the client, they share compressed statistical data for each client. The server then aggregates this data using the federated averaging (FedAvg) algorithm described in \cite{mcmahan2017communication}. It is important to note that assuming training can be completed in a single round is unrealistic due to constantly changing user data and unavailable samples. Additionally, sharing mean and variance may raise privacy concerns.

The paper in \cite{wazzeh2023towards} is the only work that explores split federated learning for continuous authentication in its framework components. The authors propose a cluster-based method for dividing federated learning models, but their approach only considers differences in unique labels without considering available resources. Additionally, client selection was done randomly, without considering resource capacity and dynamic availability for each round. 

Our methodology builds upon existing research, taking into account the limitations noted. We optimize client selection in each round by grouping clients according to their resource capacity. We employ a machine learning algorithm to predict resources and an advanced heuristic algorithm that dynamically considers resource availability. As a result, we address previous works' limitations while improving the authentication process's efficiency and security.

\section{Methodology}

\begin{figure}[H]
\centering
{\includegraphics[width=.99\textwidth]{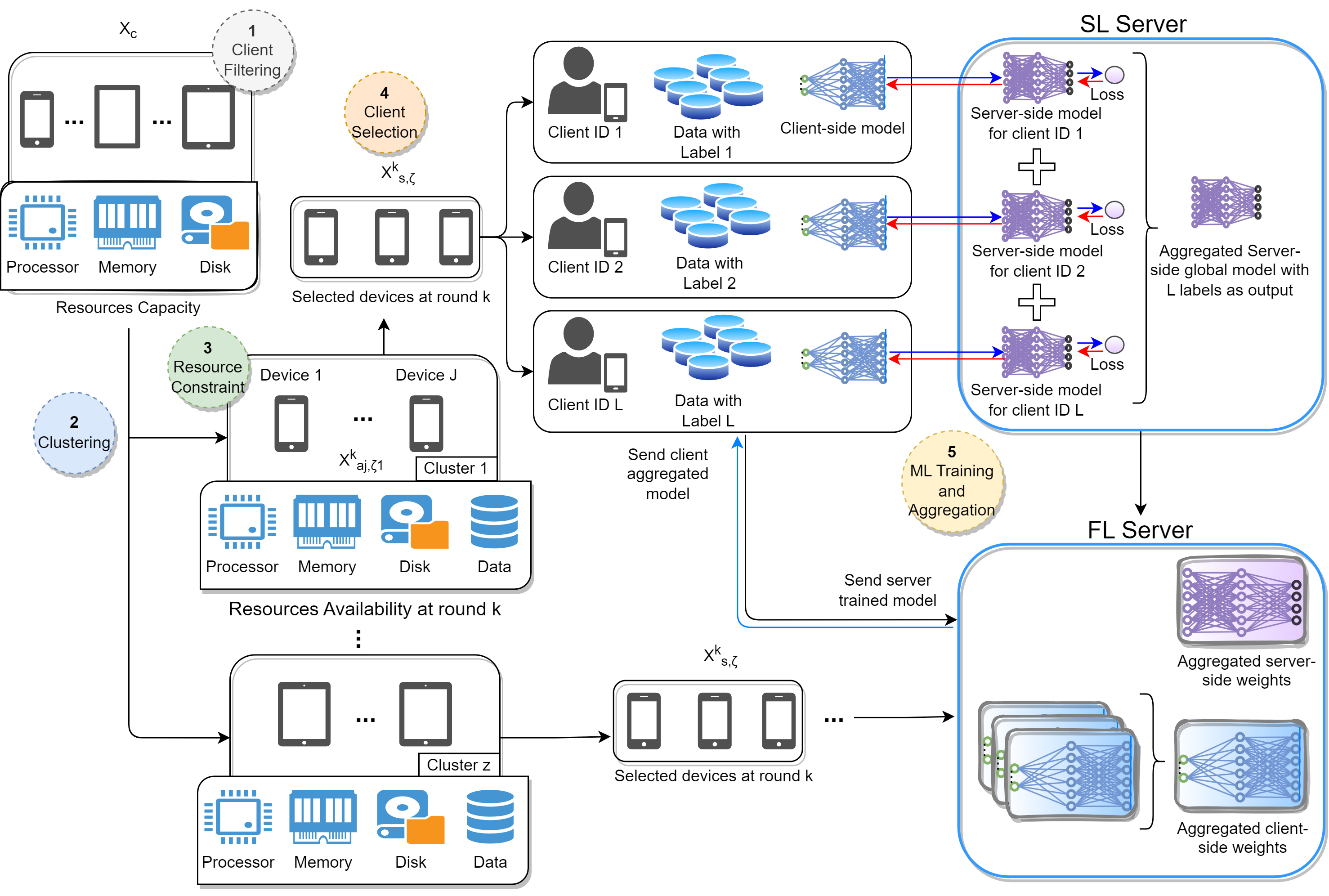}}\hfill
\caption{Overall Cluster-based Resource-aware Split Federated Learning (CRSFL) Architecture.}
\label{fig:main_architecture}
\end{figure}

\subsection{Overview}
We have developed an innovative approach to ensure continuous user authentication using split and federated learning on mobile and IoT devices. Our technique prioritizes safeguarding user privacy and model security while utilizing the benefits of distributed computing. Figure \ref{fig:main_architecture} illustrates the architecture of our proposed framework. The architecture displays a list of clients and their resources, filtering out devices that cannot train the ML model and grouping them into clusters based on resource similarity. A predictive machine learning model is then trained using rounds history and utilized in the training rounds to select resources based on dynamically available resources. Next, a heuristic approach is employed to choose clients from the list of available devices. These clients then train the model in parallel using the split server while the server part's model aggregation occurs on the split server. The client model aggregation is carried out on the federated server, and the resulting globally aggregated models are subsequently transferred to the next cluster. The notable difference between conventional FL techniques and FL tailored for continuous authentication is due to the nature of client data in such scenarios. In traditional FL, clients may have shared labels or overlapping data, unlike in continuous authentication, where each client has a distinct label, presenting unique challenges in model convergence and performance optimization. This distinction highlights the crucial role of client selection in achieving optimal continuous authentication performance.
In the following subsection, we will explain each of the architectural elements and their respective functions in more detail.

\subsection{Architectural Stages}

Our proposed approach, CRSFL, integrates various components for training machine learning models. These include clients, clusters, split learning servers, and federated learning servers, which work collaboratively in layers or stages to select clients and conduct collaborative training effectively. These encompass client filtering, clustering, resource constraints, client selection, and ML training and aggregation. By thoroughly examining each aspect, we can comprehensively describe the stages in preparing for client selection within split federated learning scenarios.

\subsubsection{Client Filtering} 
Each individual in the system is identifiable through a device with a distinct collection of data samples. Although a user may possess multiple devices in the authentication system, each is uniquely labelled. Nonetheless, all devices linked to the same user share a common label, such as the user's behavioural data. On the other hand, the labels for dissimilar users are always distinct, with each individual's behavioural data being unique to their characteristics. During the initial phase, we assess potential users to determine if they have the necessary resources to support the model. This involves comparing each user's resource capacity to the hosting requirements of the model. We assume each user has a fixed resource capacity that remains constant over time. Users who cannot meet the criteria are eliminated from the pool of potential participants. This approach ensures that only users with sufficient resources are considered for further selection.

\subsubsection{Clustering} Clusters are groups of clients managed by a split learning server. A cluster can consist of one or more devices per client and devices with similar resource capabilities are grouped to optimize performance. We employ a k-means clustering technique that can handle various devices. The k-means algorithm starts by randomly initializing cluster centroids and then iteratively updating them until convergence. At each iteration, it assigns data points to the nearest centroid based on Euclidean distance and subsequently updates each centroid to reflect the mean of the data points assigned to it \cite{sinaga2020unsupervised}. The clustering process involves sorting filtered clients into clusters based on their resource capacities, aiming to create sets of devices with similar training capabilities. This approach enhances the efficiency of split federated learning by minimizing communication overhead and enabling clients in the same cluster to collaborate on training with minimal differences in their resource capabilities. The objective is to cluster clients so that their resource profiles closely resemble each other. Doing this can reduce training idle time between devices and overall latency in a parallel split learning system. This process partitions devices into distinct groups, allowing devices within each group to train their device-side models simultaneously. This method is more effective than the vanilla split learning approach, which trains devices sequentially \cite{vepakomma2018split}. Clustering also addresses the straggler effect of device heterogeneity and network dynamics by grouping devices with matching computing capabilities \cite{tu2021feddl}.

\subsubsection{Resource Constraint} We systematically investigate each cluster to identify clients who may struggle with completing the model training process. To achieve this, we employ a Random Forest resource prediction model that draws on historical data from previous client training engagements. The Random Forest regression model is a highly effective machine-learning algorithm that utilizes decision trees to identify complex patterns within data and make predictions about the values of a target variable. The algorithm minimizes the mean squared error (MSE) between predicted and actual values to ensure precise predictions \cite{segal2004machine}. Resource constraints must be considered when training a machine learning model on devices with limited resources \cite{li2021fedmask}.
Additionally, when deploying machine learning models across multiple devices, it is essential to consider the varying resources available on each device and the differing resource demands of the model. This may result in differing model requirements among clients depending on the size of their data samples and available resources \cite{abdulrahman2020fedmccs}. To overcome this challenge, we developed a machine learning model that uses a client's resource usage history from multiple training rounds to predict actual resource utilization. We gather usage data from devices with varying resource capacities to generate the input data for our prediction model. The prediction model enables us to identify clients who may experience dropout due to resource limitations, allowing us to filter out such clients from the available pool of candidates for model training.

\subsubsection{Client Selection} The client selection process involves identifying the optimal clients for training our machine learning model. We ensure that only clients with sufficient resources to complete the training are forwarded to our heuristic selector algorithm, thus avoiding any early dropouts \cite{chahoud2023demand}. The heuristic selector algorithm then identifies the most appropriate clients with predetermined objectives. For more information on the algorithm and its objectives, please refer to section \ref{sec:problem_definition_formulation}.

\subsubsection{ML Training and Aggregation} This approach trains a multi-class classification neural network model to classify input data into different classes, each representing a specific client label. In split federated learning, each selected client device trains its model using raw data up to the cut layer. The weights obtained from this layer are then sent to the split server to complete the training process through forward and backward propagation. After the client-model weights are updated, they are transmitted to the federated server for aggregation. In a split learning architecture, the split learning server receives the server-side portion of the neural network model and the cut layer's weights or smashed data from clients concurrently. The server then trains the model's remaining layers in parallel for each device. Once the model training is complete, the server aggregates the clients' server-side model weights by averaging and transmits the final aggregated server-side weights to the federated server. A FL server initializes neural network models on a server, clusters clients, and aggregates their weights by averaging. After this, the updated weights of both the clients and the server are transferred to the next cluster to continue the training process.

\section{Problem Definition and Formulation for Client Selection}
\label{sec:problem_definition_formulation}
This section presents the problem definition and formulation for our proposed approach. We outline the mathematical formulation for the input, output, and client selection objectives. Additionally, we explain the client selection algorithm used and discuss its complexity.

\subsection{Problem Definition}
In this subsection, we will focus on tackling the challenge of client selection during split federated learning rounds. 
Initially, we evaluate potential clients during the initial phase to ensure they possess the necessary resources to support the model. We assume each client has a fixed resource capacity that remains constant over time. From there, we filter out clients who do not meet our requirements and group the remaining clients into clusters using K-means based on their resource capacities. This allows us to create sets of devices with similar training capabilities, ultimately leading to more significant split federated learning efficiency. When deploying machine learning models across multiple devices, it is crucial to consider the resources available on each device and the varying resource demands of the model. It is also important to note that differences in available resources or the size of data samples may result in differing model requirements among clients.
Our selection process aims to select clients in split federated rounds according to specific objectives. During this phase, the chosen clients work with the split server to train the machine learning model, with selection occurring in each learning round. Our client selection objectives align with our ultimate goal of continuous client authentication. The objectives can be defined as follows:
\begin{enumerate}
	\item Maximize the number of clients' devices: Select as many clients as possible during the learning phase. 

	\item Maximize the number of unique clients: During the learning phase, select as many devices with unique labels as possible since each represents a unique user requiring authentication.

	\item Maximize the learning quality: Enhancing the model's learning quality is related to carefully selecting clients with many data samples. 

	\item Minimize the client idle time: Selecting clients with similar processing capabilities is necessary for optimizing the split learning process. 

	\item Maximize the probability of a client being selected: The fitness function is designed to maximize client selection in learning rounds. 
 
\end{enumerate}

These requirements transform the problem into a multi-objective optimization problem, akin to a multi-objective Knapsack problem.
\vspace{12pt}

\textit{Multi-objective Knapsack Problem:}
The Multi-objective Knapsack Problem (MOKP) presents a unique challenge as it requires optimizing multiple conflicting objectives simultaneously, unlike the classic Knapsack problem that focuses on a single objective \cite{lust2012multiobjective}. While the standard Knapsack problem involves selecting items with specific values and weights to maximize the total value without exceeding a weight constraint, the MOKP involves multiple knapsacks (objectives) where each item can contribute to different knapsacks with varying values and capacities.

\textbf{Theorem 1.} \textit{The client selection problem in split federated learning, formulated as a Multi-objective Knapsack Problem, is NP-hard.}

\textbf{Proof}: To prove the NP-Hardness of our problem, we perform a reduction from the Knapsack Problem, specifically to the multi-objective Knapsack problem. Given an instance of the Knapsack Problem, with items, their weights, and values. We create an instance of our multi-objective optimization problem as follows:

\begin{enumerate}
    \item \textit{Items:} The clients in our problem correspond to the items in the Knapsack Problem, where a binary variable indicates whether a client is chosen in a round within a cluster. 
    
    \item \textit{Weights:} The weights in our client selection problem represent selection cost. In our problem, this weight cost is related to whether a client has enough resources to finish a round, has a large data sample size, maximizes the number of unique labels of the selected client devices, the selection of the device reduces the total selected clients idle time and the client's selection is tracked to ensure its selection within a range of rounds.
    
    \item \textit{Values:} The characteristics of each client, such as the clients selected, their data sample size, their unique labels, their processing utilization and historical selection tracking, are used by the objective functions to determine their relevance to the fitness value, similar to the way item values are considered in the Knapsack Problem.
\end{enumerate}

Thus, our objective in the multi-objective optimization problem is to maximize selected clients' fitness while adhering to the constraints outlined later in this section. By establishing this reduction, we conclude that our client selection problem is NP-Hard.

\begin{table*}
\centering
\begin{tabular}{p{2.5cm}p{12cm}}
\hline
\textbf{Notation} & \textbf{Description} \\
\hline
\(I \in \mathbb{N}\) & Total number of devices \\
\(K \in \mathbb{N}\) & The total number of rounds \\
\(k_c \in \mathbb{N}\) & The number of clusters for the clustering model \\
\(Z \in \mathbb{N}\) & Total number of clusters \\
\(J \in \mathbb{N}\) & Number of filtered devices with dynamic available resources \\
\(X_c\) & Set of devices resources capacities \\
\(X_s\) & Set of selected clients \\
\(X_{a}\) & Set of filtered clients with dynamic available resources \\
\(X_{u}\) & Set of resources utilization estimation for devices \\
\(X_{s_l,\zeta_z}^K\) & Binary variable, 1 if client \(l\) is chosen in round \(K\), 0 otherwise \\
\(\zeta\) & Set of clusters \\ 
$\bar{X}$ & The average processing utilization of selected clients \\
$W_{fq}$ & Weight associated with objective function of index $q$ \\
$F(x)$ & Multi-objective optimization function \\
$\bar{\mathbf{w}}_{a_L,\zeta_z}^{d,k}$ & Global aggregated weights at cluster $z$ for client-side at round $k$ \\
$\bar{\mathbf{w}}_{a_L,\zeta_z}^{s,k}$ & Global aggregated weights at cluster $z$ for split server-side at round $k$ \\
$G_{a_L,\zeta_z}$ & Represents the filtered clients with available resources at cluster $\zeta_z$ at round $k$ \\
$\Theta^{\in(mem, pro, dis)}$ & Represent the prediction model of resource utilization \\

\hline
\end{tabular}
\caption{Frequently Used Notations.}
\label{tab:notations}
\end{table*}

\subsection{Problem Formulation}

In this subsection, we will mathematically formulate our problem by discussing the input and output matrices to achieve an optimized client selection process and outline its objectives.

\vspace{12pt}
\subsubsection{Input}
Let \( I \in \mathbb{N} \) denote the total number of devices with different capacities, and \( X_c = (X_{c_1}, X_{c_2}, \ldots, X_{c_I}) \) represent a set of client devices comprising \(I\) smartphones and IoT devices that have resources capacity features. The client-side model training process requires careful alignment of resource utilization with the model's specific requirements to run the model on the client side regardless of the data samples they possess. To maintain a high standard of round participation, it is crucial to eliminate clients with insufficient resources from the set \(I\). As a result, the remaining set of clients, denoted as \(J\), can be represented by:

\begin{equation}
      J = \{i \in I : X_{c_i}^{\text{mem}} \geq M_{c_i}^{\text{mem}}, X_{c_i}^{\text{pro}} \geq M_{c_i}^{\text{pro}}, X_{c_i}^{\text{dis}} \geq M_{c_i}^{\text{dis}}\}
\end{equation}

\begin{description}[itemsep=3pt, labelwidth=0.9cm, align=left]
    \item[\(X_{c_i}^{\text{mem}}\)] : Memory capacity of the client device.
    \item[\(X_{c_i}^{\text{pro}}\)] : Processing unit capacity of client device.
    \item[\(X_{c_i}^{\text{dis}}\)] : Hard disk capacity of client device.
    \item[\(M_{c_i}^{\text{mem}}\)] : Memory loading requirement of the model.
    \item[\(M_{c_i}^{\text{pro}}\)] : Processing unit loading requirement of the model.
    \item[\(M_{c_i}^{\text{dis}}\)] : Hard disk loading requirement of the model.
\end{description}

\vspace{12pt}

The filtered client set \( J \) serves as input for the clustering component. Let \( \zeta = \{\zeta_1, \zeta_2, \ldots, \zeta_Z\} \) denote the set of clusters, where \( Z \) is the total number of clusters, and \( X_{c_j,\zeta_z} \) represents client \( j \) in cluster \( z \).

Let us define $X_a$ as a set of filtered clients with dynamically available resources. The challenge is to train a Random Forest model $M$ on distributed, resource-constrained devices while accounting for varying resource demands and available data sample size(\(X_{a_j,\zeta_z}^{\text{sam}, k}\)). We assume that the data sample size remains unchanged between the rounds. Moreover, the machine learning model's resource demands may differ across clients due to differences in input data requirements. Let $K$ be the total number of rounds for training a machine learning model. Let us define \( X^k_{a,\zeta_z}\) as the set of clients with dynamically available resources at round \(k\) in a cluster \(z\). 

Denote $X_{u_j,\zeta_z}^k$ as the estimated resource utilization of memory $X_{u_j}^{\text{mem}, k}$, processing power $X_{u_j}^{\text{proc}, k}$, and hard disk $X_{u_j}^{\text{dis}, k}$ for client $j$ at round $k$ within cluster $z$, during model training.

To guarantee consistency and quality in the model's training, it is critical to filter out clients lacking sufficient resources and include only those that meet the minimum requirements. This process involves systematically removing inadequately resourced clients from the original set, denoted as \(J\), resulting in a new set, denoted as \(L\). By doing so, we can ensure that the remaining client set comprises individuals capable of contributing meaningfully to the model training process. Subject to the following constraints:

\begin{equation}
      L = \{j \in J : X_{u_j}^{\text{mem}} \leq M_{a_j}^{\text{mem}}, X_{u_j}^{\text{pro}} \leq M_{a_j}^{\text{pro}}, X_{u_j}^{\text{dis}} \leq M_{a_j}^{\text{dis}}\}
\end{equation}

\begin{equation}
\label{eq:resource_prediction}
X_{u_j}^k = \frac{1}{N} \sum_{i=1}^{N} \text{Tree}_t(X_{a_j,\zeta_z}^k)
\end{equation}

Equation \eqref{eq:resource_prediction} establishes a relationship between observed resource use and the predicted value using a Random Forest regression model, where $N$ is the number of decision trees in the ensemble, and $\text{Tree}_t(X_{a_j,\zeta_z}^k)$ represents the prediction made by the $t$-th decision tree for the input features. The equation indicates that the final prediction $X_{u_j}^k$ is the average of predictions made by all decision trees in the Random Forest ensemble for the given input features. Finally, the input to our optimization problem is $X_{a_j,\zeta_z}^k$.

\vspace{12pt}
\subsubsection{Output}
The selection process output adheres to specific clients who can contribute effectively to the split learning task by achieving predetermined objectives. We aim to maximize the selection in each round \(k\) by choosing a set of clients \(X_{s_L,\zeta_Z}^K\), which contains binary values, where 1 represents if the client is selected and 0 otherwise. The list represents the number of compatible clients from \(L\) that adhere to the defined resource constraints and follow the objective functions.

\vspace{12pt}
\subsubsection{Client Selection Objectives}
The objective involves selecting a suitable group of clients (\(X_{s,\zeta_z}^K\)) in every round $k$ of a cluster $\zeta_z$ from a diverse pool of available clients (\(X_{a_l,\zeta_z}^k\)). Each client has a different resource profile, label and data sample size. Our objectives include the number of data sample sizes each client has (\(X_{a,\zeta_z}^{\text{sam}, k})\), the number of selected devices (\(X_{s,\zeta_z}^k\)), the number of unique labels of the selected client devices ${u}(X_{s,\zeta_z}^k)$, the differences in the processing capabilities of the selected clients $\mathbf{V}(X_{u,\zeta_z}^{proc, k})$ and finally to monitor the clients selected by tracking the historical client selected ${hist}(X_{s,\zeta_z}^k)$ so that no client device is left out of the training process. Clients are chosen based on their varying resource limitations, and their security goals are maintained. Each client is assigned unique labels and is responsible for optimizing their participation while ensuring the continuous authentication model's maximum performance. Table \ref{tab:notations} defines the summary of some important notations used.

We introduced a weighting mechanism based on the method of adjustable weights \cite{pardalos2012pareto} to enhance flexibility in prioritizing different objective functions. These weights are decimal values ranging from 0 to 1; their sum always equals 1. This combination is expressed as follows:

\begin{equation}
W_{f1} + W_{f2} + W_{f3} + W_{f4} + W_{f5} = 1
\end{equation}

This allows us to assign varying importance to each objective while ensuring a balanced overall priority \cite{sami2020dynamic}. The clients selected need to adhere to the following objective functions:\\

\begin{enumerate}

\item Maximize the number of clients' devices:
\begin{equation}
f_1 = \max \left(\left( \sum_{l=1}^{L} X_{s_l,\zeta_z}^k \right) \cdot W_{f1} \right)
\end{equation}

\begin{itemize}
  \item \(\max_{X_{s,\zeta_z}^k}\) indicates that the maximization is performed over the set of selected clients in round \(k\), within a cluster $\zeta_z$, denoted by \(X_{s,\zeta_z}^k\).
  \item \(\sum_{l=1}^{L} X_{s_l,\zeta_z}^k \cdot W_{f1}\) represents the sum over all selected clients (\(l=1\) to \(L\)) in round \(l\).
  \item \(X_{s,\zeta_z}^k\) is a set of binary variables indicating whether a client is selected or not in round \(k\) within a cluster $z$.
  \item \(W_{f1}\) is the weight associated with the entire summation, indicating the importance of selecting clients.
\end{itemize}
This selection is necessary to enable the model to consider the differences in client features and labels accurately. Therefore, having more selected clients can help speed up the model's process, as the data sample size between the client devices will be more significant. Clients' active participation in the learning phase contributes to a quick convergence of the model.

\item Maximize the number of unique clients:

\begin{equation}
f_2 = \max \left(\left( \sum_{l=1}^{L} {u}(X_{s_l,\zeta_z}^k) \right) \cdot W_{f2} \right)
\end{equation}

\begin{itemize}
  \item \(\max_{X_{s,\zeta_z}^k}\) indicates that the maximization is performed over the set of selected clients in round \(k\), within a cluster $\zeta_z$, denoted by \(X_{s,\zeta_z}^k\).
  \item \(\left(\left( \sum_{l=1}^{L} {u}(X_{s_l,\zeta_z}^k) \right) \cdot W_{f2} \right)\) represents the sum over all selected unique clients (\(l=1\) to \(L\)) in round \(k\).
  \item \(X_{s,\zeta_z}^k\) is a set of binary variables indicating whether a client is selected or not in round \(k\) within a cluster $\zeta_z$.
  \item \(W_{f2}\) is the weight associated with the entire summation, indicating the importance of selecting clients.
\end{itemize}
Having more selected clients with different labels can help speed up the process of building the model. The active participation of clients in the learning phase ensures a thorough understanding and contributes to the swift convergence of the model. The model learned will differentiate between the different user's features.

\item Maximize the learning quality:
        
\begin{equation}
f_3 = \max \left(\left( \sum_{l=1}^{L}  X_{s_l,\zeta_z}^k \cdot X_{a_l,\zeta_z}^{sam, k} \right) \cdot W_{f3} \right)
\end{equation}

\begin{itemize}
  \item \(\sum_{l=1}^{L} \left( X_{s_l,\zeta_z}^k \cdot X_{a_l,\zeta_z}^{sam, k} \right) \cdot W_{f3}\) represents the sum over all selected clients (\(l=1\) to \(L\)) in round \(k\), where \(X_{a_l,\zeta_z}^{sam, k}\) is the sample size of client \(l\) in round \(k\) within a cluster $z$.
  \item \(W_{f3}\) is the weight associated with the entire summation, indicating the importance of selecting clients with larger sample sizes.
\end{itemize}
This ensures the model is exposed to a diverse and comprehensive set of training examples. The emphasis on selecting clients with high data samples is a strategic approach to increase the model's ability to generalize and make accurate predictions.

\item Minimize the client's idle time:
  
\begin{equation}
f_4 = \min \left( \left( \sum_{l=1}^{L} X_{s_l,\zeta_z}^k \cdot \mathbf{V}(X_{u_l,\zeta_z}^{proc, k}) \right) \cdot W_{f4} \right)
\end{equation}

\begin{itemize}
    \item $\mathbf{V}(X_{u_l,\zeta_z}^{proc, k})$: Represents the variance of the processing utilization estimates among the selected clients. The variance can be calculated as follows: $\mathbf{V}(X_{u_l,\zeta_z}^{proc, k}) = \frac{1}{L} \sum_{l=1}^{L} \left(X_{u_l,\zeta_z}^{proc, k} - \bar{X}\right)^2$, where $\bar{X}$ is the average processing utilization calculated as $\bar{X} = \frac{1}{L} \sum_{l=1}^{L} X_{u_l,\zeta_z}^{proc, k}$.
    \item \(X_{u_l,\zeta_z}^{proc, k}\): The processing utilization estimation of the model on client \(l\) in round \(k\) within a cluster $z$.
    \item $W_{f4}$: This is the weight associated with the entire objective function, indicating the importance of this objective in the overall optimization process.
\end{itemize}
This minimizes the variance of processing utilization estimates among the selected clients, encouraging the selection of clients with processing capabilities close to the average. This approach enhances the overall effectiveness of federated learning by ensuring that clients progress through rounds with minimal wait times.

\item Maximize the probability of a client being selected:

\begin{equation}
    f_5 = \max \left( \left( \sum_{l=1}^{L} X_{s_l,\zeta_z}^k \cdot \text{hist}(X_{s_l,\zeta_z}^k) \right) \cdot W_{f5} \right)
\end{equation}

\begin{itemize}
    \item $\text{hist}(X_{s_l,\zeta_z}^k)$: Historical record for client $l$ in round $k$ within cluster $\zeta_z$, where $\text{hist}(X_{s_l,\zeta_z}^k) = 1$ if client $l$ was shown but not selected, and $0$ otherwise.
    \item $W_{f5}$: The weight associated with the entire objective function, indicating the importance of this objective in the overall optimization process.
\end{itemize}
The function aims to engage a diverse range of participants who may have been overlooked. Additionally, maximizing client selection helps in improving model generalization and robustness. It encourages the participation of a broader range of data distributions and characteristics, thereby improving the overall quality of the learning rounds.
\end{enumerate}

The optimization problem can be formulated as follows:

\[
\begin{aligned}
    &\text{Maximize} \quad \sum_{l \in X_a}  X_{s_l,\zeta_z}^K \cdot W_{fq} \\
    &\text{where} \\
    &\quad X_{s_l,\zeta_z}^K \in \{0, 1\} \quad \forall l, z, K \\
    &\quad \sum_{q=1}^{Q} W_{f_q} = 1
\end{aligned}
\]

The multi-objective optimization problem is represented as follows:
\begin{equation}
 F(x) = f_1(x) + f_2(x) + f_3(x) - f_4(x) + f_5(x)
\end{equation}

Where:
\begin{align*}
f_1(x) & = \text{Number of active devices} \\
f_2(x) & = \text{Number of unique clients} \\
f_3(x) & = \text{Quality of learning} \\
f_4(x) & = \text{Reduction in device idle time}\\
f_5(x) & = \text{Prioritize new devices}
\end{align*}

\subsection{Genetic Algorithm For Client Selection Problem}
\begin{figure}[H]
\centering
{\includegraphics[width=.80\textwidth]{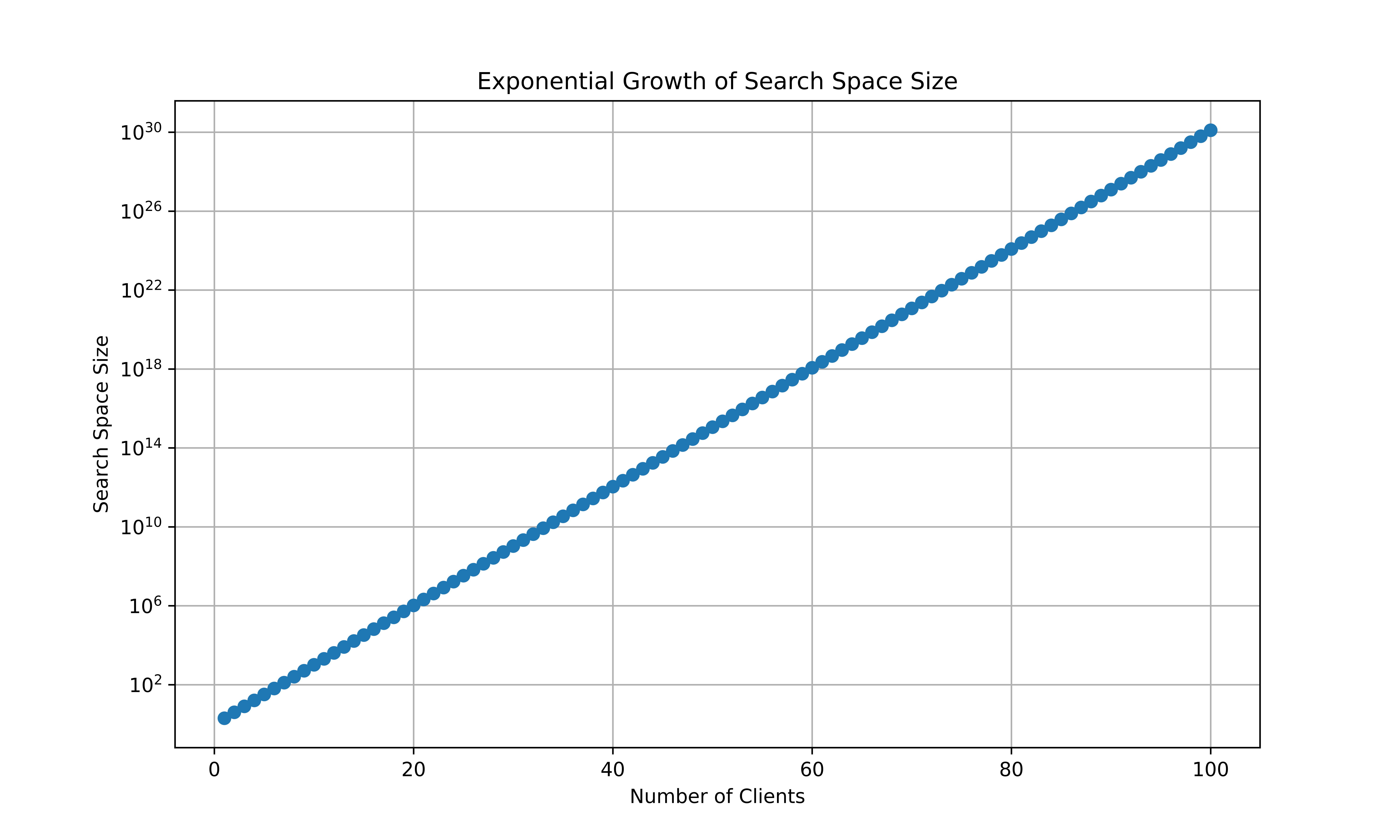}}\hfill
\caption{Search Space Size for Client Selection with Multi-Objective Optimization.}
\label{fig:search_space}
\end{figure}

\begin{figure}[H]
\centering
{\includegraphics[width=.80\textwidth]{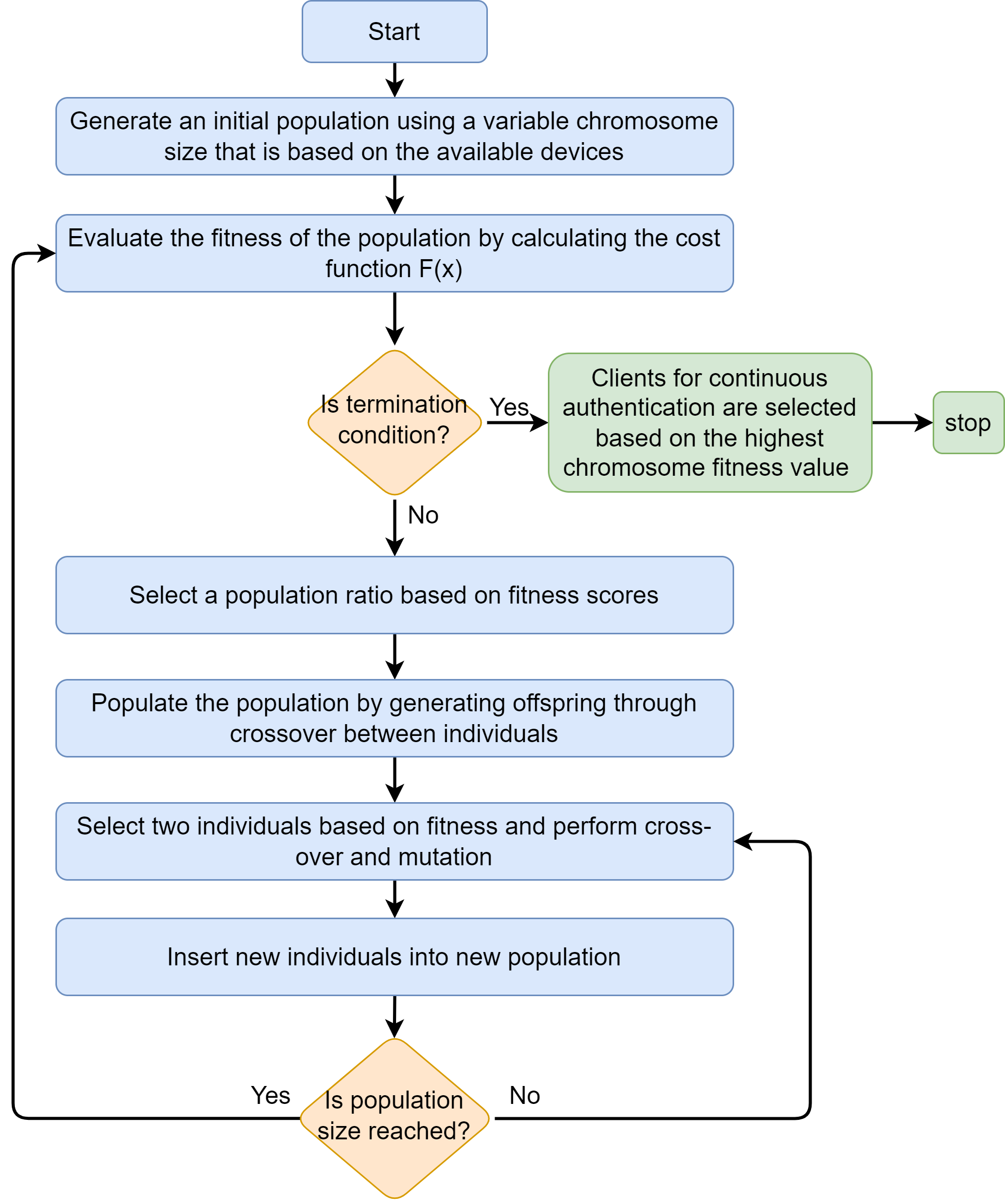}}
\caption{Genetic Algorithm Steps for Client Selection in Continuous Authentication.}
\label{fig:ga_flow}
\end{figure}

Multi-objective optimization problems inherently feature multiple solutions rather than a single optimal one, and these solutions are referred to as Pareto solutions \cite{murata1995moga}. Acquiring the Pareto set solution efficiently is crucial. A genetic algorithm (GA) proves advantageous in addressing this challenge due to its ability to explore the search space efficiently. The GA mirrors the natural selection process, favouring the reproduction of the fittest set of solutions for subsequent generations \cite{konak2006multi}. Non-GA approaches, with their expansive search spaces as reported in Figure \ref{fig:search_space}, require thorough exploration, often demanding significant computational resources and time to identify optimal solutions efficiently, as the number of possible solutions is exponential to the increased number of clients. Each chromosome is represented as a \(X_{s_{\zeta_z}}^k\) matrix and signifies the decision made by the optimization model regarding the selection of the client for model training. Mathematically, \(X_{s_l}^k \in [0, 1]\). This representation captures the binary nature of the decision, facilitating the optimization process within the GA framework. Figure \ref{fig:ga_flow} gives an overview of the genetic process. It starts with the initialization of the population and goes through the various stages of selection, Crossover, Mutation, and fitness evaluation. This cycle is repeated iteratively to carry out the GA operations.

To efficiently solve a mixed-integer programming (MIP) problem, it is essential to analyze its time complexity and break it into various components. By doing so, we can identify the most computationally intensive parts of the problem.

In the \textit{Initialization} phase, we generate a group of potential solutions (chromosomes) based on the total number of clients available for training. Our first step is to create an initial population of chromosomes utilizing a set of genes. Each gene is tailored to the specific problem we are addressing. In our case, a gene represents a client model within a chromosome encompassing a subset of clients. To create a chromosome, we randomly select genes, ensuring no duplicate genes are within the same chromosome.

During The \textit{Selection} process, individuals are chosen from the population to become parents for the next generation. To filter the population, we select the best half before proceeding with Crossover and Mutation to create the new generation.

\textit{Crossover} is a process that combines genetic material from two parent solutions to produce new offspring solutions. During Crossover, two chromosomes are randomly split at a position, and their segments are swapped when a randomly generated float number between 0 and 1 exceeds a predefined crossover ratio of 0.5 specified by the GA configuration.

\textit{Mutation} is a process that introduces random changes in offspring solutions to maintain diversity. Offspring solutions replace some individuals in the current population to form the next generation.

The algorithm's \textit{Termination} criteria determine when to halt the optimization process. These criteria are based on factors such as reaching a maximum fitness threshold or completing a specified number of generations. Genetic selection results in a chromosome that includes clients whose combined selection meets the predefined objectives.

To measure the complexity of the optimization process, we use \(N\) to represent the total number of decision variables, \(G\) to represent the number of Generations in the optimization process, \(L\) to denote the number of clients selected at each round, and \(Q\) to signify the number of objective functions. This complexity is comprised of various components, including:

\begin{description}
    \item[$C_{\text{initialization}}$] \hfill \\
    Complexity of the initialization step: \(O(N \times L )\).
    
    \item[$C_{\text{selection}}$] \hfill \\
    Selection process complexity: \(O(N)\).
    
    \item[$C_{\text{crossover}}$] \hfill \\
    Crossover complexity over rounds: \(O(N \times K)\).
    
    \item[$C_{\text{mutation}}$] \hfill \\
    Mutation complexity over rounds: \(O(N \times K)\).
    
    \item[$C_{\text{fitness}}$] \hfill \\
    Complexity of calculating the multi-objective optimization function: \(O(N \times J \times Q)\)
\end{description}

Thus, the overall time complexity can be expressed as \(O(G \times (C_{\text{initialization}} + C_{\text{selection}} + C_{\text{crossover}} + C_{\text{mutation}} + C_{\text{fitness}}))\).

\section{CRSFL Framework}
The process of our proposed architecture lays out the specific steps and procedures we use to ensure that the neural network model is trained across different clusters of clients. The method outlined in the prior section is followed to process the input of Algorithm \ref{alg:main_alg}. This involves carrying out client filtering, model prediction, and K-means clustering. Next, we will describe the presented pseudo-code for the proposed framework:

\begin{algorithm}
\DontPrintSemicolon

    \caption{Proposed Clustered-based Resource-aware Split Federated Learning (CRSFL) Algorithm.}
    \label{alg:main_alg}

    \KwIn{$X_{a,\zeta_Z},\Theta^{\in(\text{mem}, \text{pro}, \text{dis})}, K$} 
    \KwOut{$\bar{\mathbf{w}}_{a_J,\zeta_z}^{d,k}$, $\bar{\mathbf{w}}_{a_J,\zeta_z}^{s,k}$} 
    \BlankLine
    
    \For{$k$ in $K$}{
        \For{$X_{a,\zeta_z}^k$ in $X_{a,\zeta_Z}^k$}{ 
            $L = 0$ \;
            collect $X_{a_j,\zeta_z}^{\text{mem}, k} \land X_{a,\zeta_z}^{\text{pro}, k} \land X_{a_j,\zeta_z}^{\text{dis}, k} \land X_{a_j,\zeta_z}^{\text{sam}, k}$\;
  
            $X_{u_j,\zeta_z}^{\in(\text{mem}, \text{pro}, \text{dis}), k} = \Theta^{\in(\text{mem}, \text{pro}, \text{dis})} =X_{a_j,\zeta_z}^{\in(\text{mem}, \text{pro}, \text{dis}), k}$ \;
        
           \If{$X_{a_j,\zeta_z}^{\text{mem}, k} \geq X_{u_j,\zeta_z}^{\text{mem}, k} \land X_{a_j,\zeta_z}^{\text{pro}, k} \geq  X_{u_j,\zeta_z}^{\text{pro}, k} \land  X_{a_j,\zeta_z}^{\text{dis}, k}  \geq X_{u_j,\zeta_z}^{\text{dis}, k}$} 
           {
                $L += 1$ \;
                add $X_{a_j,\zeta_z}^k$ to $G_{a_L,\zeta_z}$ \;
            }

        \BlankLine
        $X_{s,\zeta_z}^k$ = \text{ga\_selector}($G_{a,\zeta_z}^k$)\;
        \BlankLine
        \For{$X_{s_l,\zeta_z}^k$ in $X_{s_L,\zeta_z}^k$}{
            \tcp{\small Client loads device-side global model}
            $(X_{s_l,\zeta_z}^k).\text{load\_state\_dict}(\bar{\mathbf{w}}_{a_L,\zeta_z}^{d,k})$\;
            
            \tcp{\small server loads server-side global model}
            \text{server\_weights}($X_{s_l,\zeta_z}^k$).\text{load\_state\_dict}$(\bar{\mathbf{w}}_{a_L,\zeta_z}^{d,k})$\;

            \For{$e$ in range(epochs) }{
                \text{client\_model\_output} = $M_{X_{s_l,\zeta_z}}^{d,k}$\text{(client\_data)}\;
                \text{smashed\_data} = \text{client\_model\_output.requires\_grad\_(True)}\;
                \text{server\_model\_output} = $M_{X_{s_l,\zeta_z}}^{s,k}$\text{(smashed\_data)}\;
                \text{loss} = \text{criterion(server\_model\_output, labels)}\;
                \text{loss.backward()}\;
                \text{client\_grad} = \text{smashed\_data.grad}\;
                $M_{X_{s_l,\zeta_z}}^{d,k}$.\text{backward(client\_grad)}\;
            }
                send $M_{X_{s_l,\zeta_z}}^{d,k}$\text{.state\_dict() to fed\_server}\;
        }
        
        \tcp{\small Fed\_Server updates global\_client\_aggregated\_weights}
        $\bar{\mathbf{w}}_{s_L,\zeta_z}^{d,k} = \frac{1}{L} \sum_{l=0}^{L} (X_{s_l,\zeta_z}^k \mathbf{w}_{s_l,\zeta_z}^{d,k})$
        
        \tcp{\small Split\_Server updates global\_server\_aggregated\_weights}
        $\bar{\mathbf{w}}_{s_L,\zeta_z}^{s,k} = \frac{1}{L} \sum_{l=0}^{L} (X_{s_l,\zeta_z}^k \mathbf{w}_{s_l,\zeta_z}^{s,k})$
        }
    }
\end{algorithm}

\begin{enumerate}
\item \textit{Client Selection for training:} To begin training, we select the initial cluster of clients with the highest resource capacities. This ensures that the initial training is done with the most capable clients and that the dropout rate will be minimal compared to other clusters. We evaluate each client selected with available resources and data samples to train their respective client-side model. The split learning server then requests client information about their available resources and the number of data samples. Algorithm \ref{alg:main_alg} refers to this step in lines 1 $\rightarrow$ 8. Upon receiving this information, the server selects a combination of clients to maximize the fitness value collectively (line 9). This selection of clients is an NP-hard problem, as previously explained. Therefore, a heuristic approach is employed on the server for client selection, using a genetic algorithm to choose the best-suited clients to achieve the objectives set by the server. The selected participants then receive the global client-side neural network model from the federated server and train each on their data in parallel.

\item \textit{Parallel Client-side Training:} Client clusters use a collaborative training approach where each client trains their model layers simultaneously, transmitting only the gradients of the cut layer. This is accomplished by processing individual sample data up to the cut layer of the neural network model through a forward pass. However, marginal errors may impact clients, such as dropping out due to a sudden increase in resource usage \cite{li2022smartphone}. Resource fluctuations could prevent clients from transmitting updated gradient information to the split server or cause them to go offline and miss receiving gradient information from the server to update their weights (lines 10 $\rightarrow$ 15).  

\item \textit{Parallel Server-side Training:} In parallel, the split learning server simultaneously receives updates to the weights from the client side, trains the remaining layers of the model separately for each client, and continues the forward pass initiated by the client. The server calculates the model output to determine the learning loss rate once the forward pass is finished. After evaluating the loss, the split learning server performs back-propagation through the remaining layers to calculate gradients for the parameters in those layers. The server then transmits these updated gradients to the corresponding clients, enabling them to modify their local parameters and proceed with the training process (lines 16 $\rightarrow$ 21).

\item \textit{Weights Aggregation:} After undergoing multiple epochs of split learning training, the split learning process now involves a matrix that includes the client ID device and its server-side gradients. The split server then aggregates these weights to create a new global server-side model for the cluster. Additionally, the federated server acts as an aggregator for the client-side models, which may be the same split server or a new server component to enhance privacy and prevent complete model weights from being held by a single entity. The federated server performs a weighted aggregation for the client-side model to create a new global model for the same cluster. Finally, the Federated and Split servers send the aggregated client-side model to the clients in the subsequent cluster and use the aggregated server-side model for the upcoming split learning training. This guarantees that the clients in the following cluster will have some pre-trained model weight for both parts of the model (lines 22 $\rightarrow$ 23).
\end{enumerate}

\section{Experimental Setup}

\begin{figure}[H]
\centering
{\includegraphics[width=.80\textwidth]{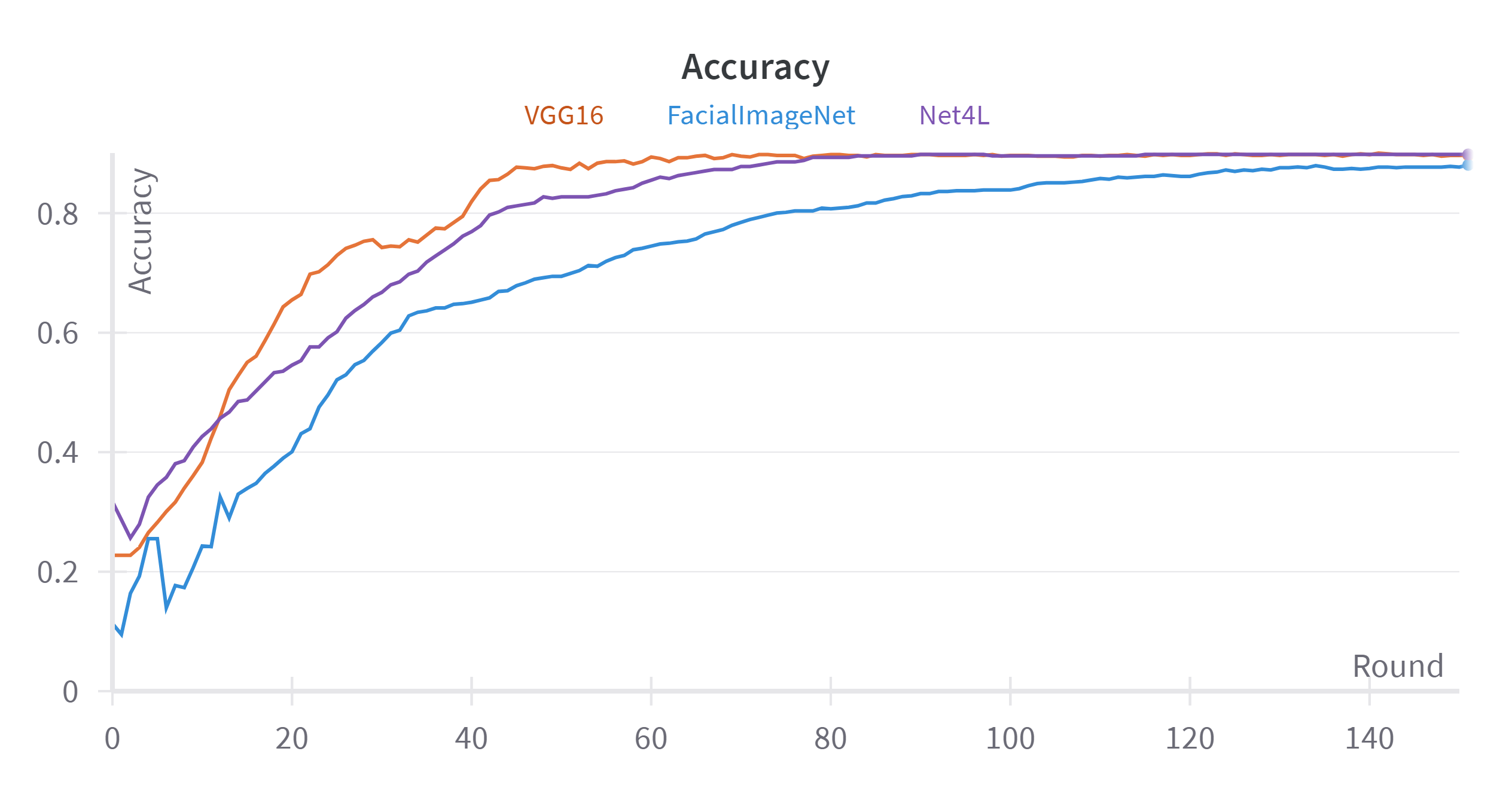}}\hfill
\caption{Accuracy Metric Performance for Different Machine Learning Models.}
\label{fig:accuracy_benchmark}
\end{figure}

In the following section, we will provide details on our datasets, experimental setup, the experiments we performed, and the results we obtained from these experiments.

We conducted the experiments using the modular federated learning framework outlined in \cite{arafeh2023modularfed}. In a Windows environment, our experiments were performed on  \href{https://pytorch.org/}{$Pytorch^1$}. Furthermore,  \href{https://www.docker.com/}{$Docker^2$} containers were employed to simulate the cluster resources profile, with memory being restricted to 2GB for every container and varying CPU capacities. We utilized a machine equipped with eight-core processors with a Base speed of 3.59GHz and 32GB of RAM, and we employed different CPU counts of 1.5, 2, and 2.5 as CPU capacity for devices.
We customized the configuration for our authentication use case and tested three different multi-class classification machine learning models in which we show the benchmark results in Figure \ref{fig:accuracy_benchmark}. The figure shows a VGG16 model with 16 layers. In comparison, the FacialImageNet model uses five layers with two convolutional layers. Finally, the Net4L contains four layers that produce comparable levels of accuracy as the VGG16 model while requiring less processing power. The model requires 128*128 as input and 128*3 as hidden layers. For simplicity, we divided the model in half for the client and server for split learning. Following a thorough hyperparameter configuration process, we identified the optimal settings for our experiments. Our chosen optimizer was stochastic gradient descent, with a fixed learning rate of 0.01 and a momentum of 0.9. We also utilized the cross-entropy loss function. The data split ratio was set to 8:2 for training and testing. We referred to the work of \cite{li2023convergence} when selecting the configuration for neural network training. In addition, we visualized the resulting plots using Weights \& Biases framework \cite{wandb}.

\subsection{Dataset}
We conducted various experiments to evaluate the performance of our proposed approach for continuous authentication using a real-life facial authentication dataset.

\begin{figure}[htp]
\centering
{\includegraphics[width=.80\textwidth]{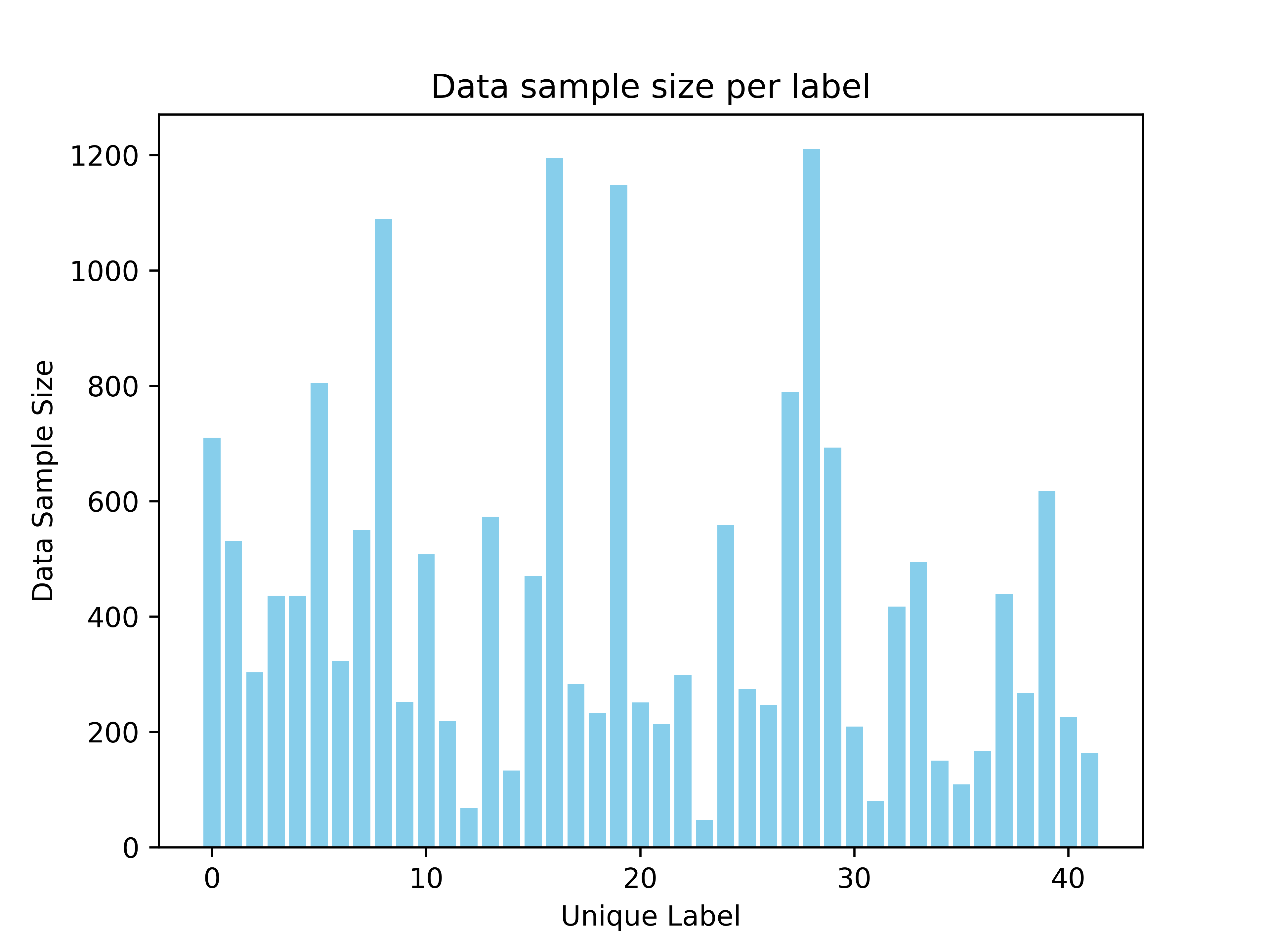}}\hfill
\caption{Data Sample Size per Client in UMDAA02-FD Dataset.}
\label{fig:dataset_details}
\end{figure}

\textbf{UMDAA-02-FD Filtered:} 
The UMDAA-02-FD Face Detection dataset \cite{mahbub2016active} presents a range of challenges for vision algorithms due to its diverse range of images, which includes partially visible faces, various lighting conditions, occlusions, and facial emotions. The dataset comprises 33,209 photos taken at 7-second intervals from 44 users, making it an ideal choice for continuous authentication scenarios. We processed the dataset to a fixed size of 128x128 to reduce the computational burden. We created a custom dataset from the original data of UMDAA-02-FD to hold the valid user training samples. The new dataset is processed through several cleaning steps, ensuring that only the valid training images are considered for training; we removed two users from the initial set as they were sharing some familiar facial photos between them. We perform the following steps to process the data: 
\begin{enumerate}
    \item  A sample of images falling below a specified brightness threshold was initially removed. We experimented with various threshold values for the brightness (ranging from 5 to 25), ultimately setting 15 as the threshold criterion. 
    \item We employed the Laplacian variance method with various predefined thresholds, determining that a threshold of 15 most effectively-identified and eliminated blurred images \cite{bansal2016blur}.
    \item We converted images to gray-scale by applying data normalization using the minimum and maximum values of the input data followed by histogram equalization using the tools from \href{https://opencv.org/}{$OpenCV^3$}.
    \item The images were then resized to 128x128 pixels for more efficient machine-learning training. It is worth noting that these preprocessing steps led to the removal of nearly 7\% of inadequate training samples from the dataset and two users with shared data, thereby enhancing its overall quality and suitability for subsequent analysis.
\end{enumerate} 
For our experiments, we tested using multiple dynamic configurations while considering that for each device, a portion of the user data will be available, such as for each unique client label ranging from 0 to 41, a random number of devices (2-6) will be assigned to his data. The number of data samples per device differs from one user to another since the data is non-IID by default; we used a data generator for each device to randomly select data points ranging between 100 and 150, thus maintaining non-IID characteristics while constraining the range. The user's data distribution of the dataset is visualized in Figure \ref{fig:dataset_details}; the bar chart shows the total number of data samples per label, in which the unique case of non-IID case exists. 

\subsection{Experiments}

Our experiments show different configurations when training a machine-learning model. We focus on accuracy by comparing the centralized experiment, in addition to metrics related to the number of dropping clients, the training time, traffic required, and total idle time.

\textbf{Centralized Training (Cen):} The concept of centralized experiments is widely used in machine learning. It involves transmitting the raw data to a server, where a centralized model is trained on the entire data.

\textbf{Split Learning (SL):} Split learning enables private machine learning by dividing the model between the client and server. This way, client data remains secure, and only model weights are transmitted during the local data training process. The experiment is performed sequentially, meaning the selected clients must wait for each other to complete the training before starting. To ensure a consistent training latency, we allocated a dedicated training latency budget exclusively for this experiment, matching the CRSFL experiment.

\textbf{Clustered Split Federated Learning (CSFL):} We applied a clustered split federated learning algorithm as reported in \cite{wazzeh2023towards}. The approach does not consider client selection optimization; therefore, the clients are randomly selected in every cluster.

\textbf{Clustered Resources Split Federated Learning (CRSFL):} We applied our proposed approach in this experiment, where the clients are selected from every cluster in every round based on the availability of the resources they each have. Client selection is optimized by using a heuristic GA approach, which is based on some predefined objectives.

\subsection{Results}

\begin{figure}[htbp]
\centering
{\includegraphics[width=.80\textwidth]{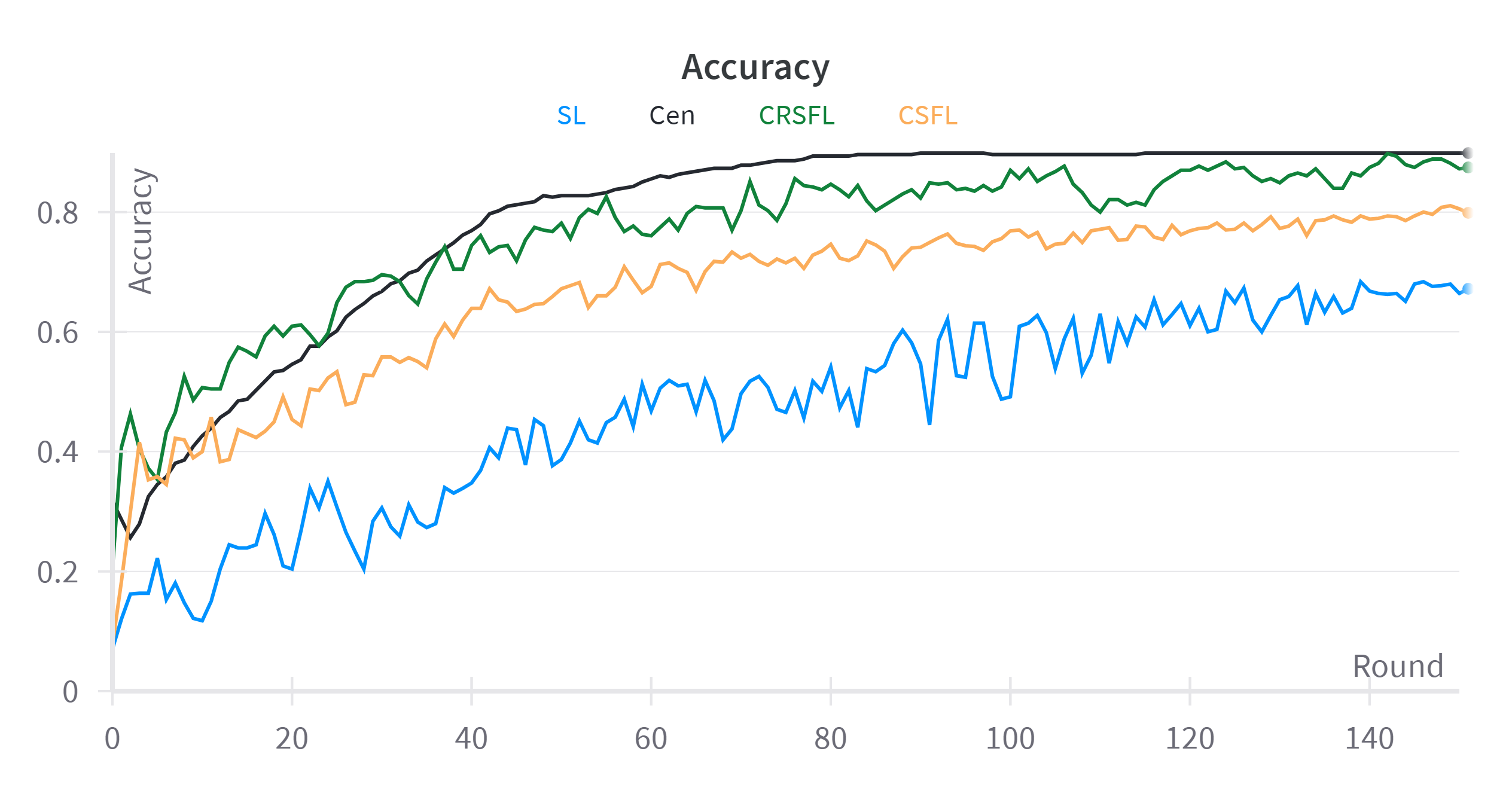}}\hfill
\caption{Test Set Accuracy per Round.}
\label{fig:accuracy}
\end{figure}

\begin{figure}[htbp]
\centering
{\includegraphics[width=.80\textwidth]{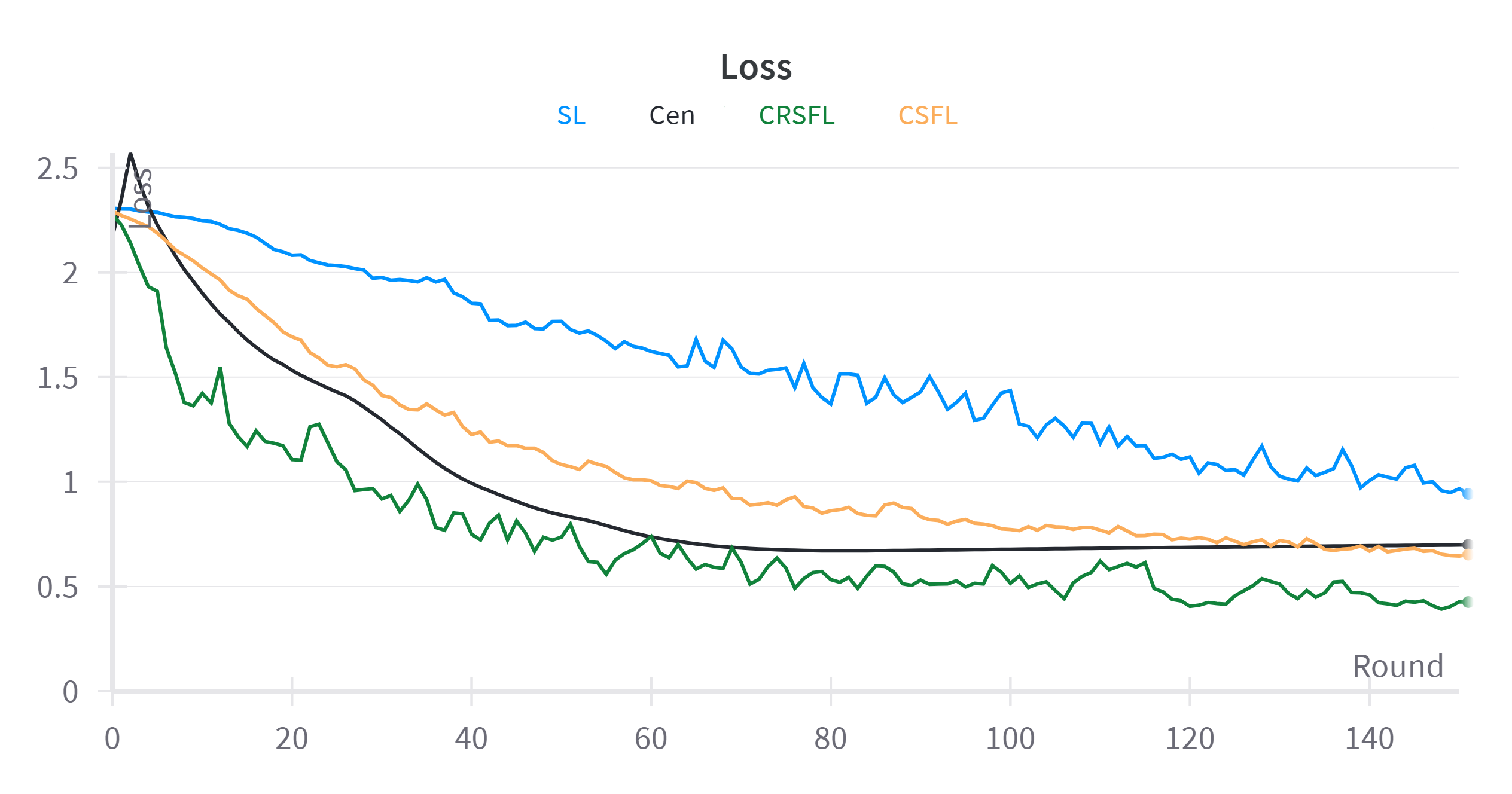}}\hfill
\caption{Test Set Loss per Round.}
\label{fig:loss}
\end{figure}

\subsubsection{Performance Metrics: Accuracy and Loss}
Figures \ref{fig:accuracy} and \ref{fig:loss} show the accuracy test results and loss for UMDAA02-FD over communication rounds. 
Starting with the centralized approach labelled "Cen," the model's convergence has reached an impressive 90\% accuracy despite the dataset's significant challenges. However, our proposed approach, "CRSFL," which utilizes GA selection, achieves a similar level of accuracy at 89\% after 150 rounds. In contrast, the random selection approach, "CSFL," progresses steadily but remains at 80\% after 150 rounds. Finally, the split learning approach represented by "SL" slowly converges and only reaches 65\% at the end of training. The slower convergence of SL and CSFL is due to their inefficient client selection processes and failed resource constraints resulting from random selection. 

\begin{figure}[H]
\centering
{\includegraphics[width=.80\textwidth]{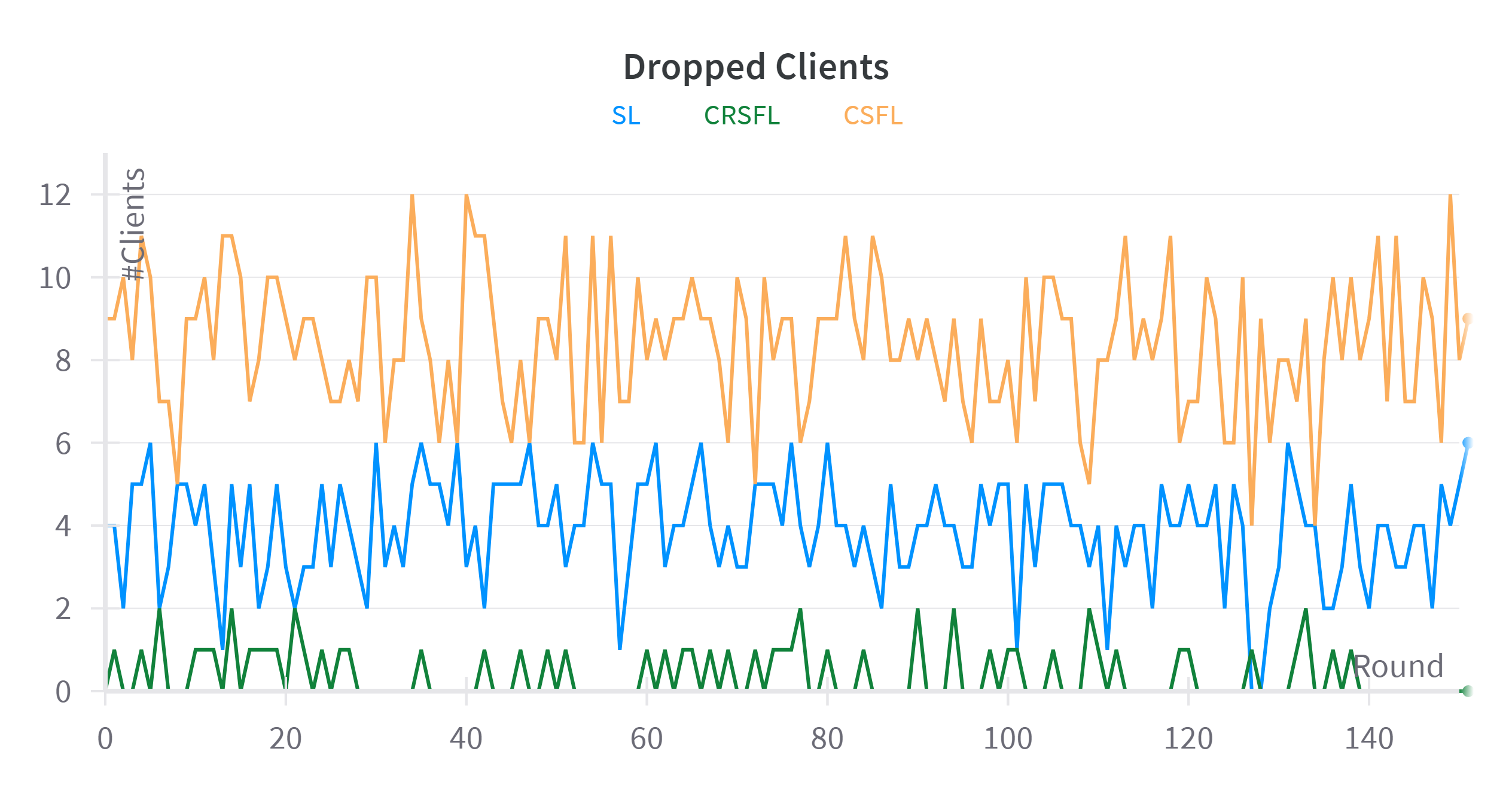}}\hfill
\caption{Total Number of Dropped Clients per Round.}
\label{fig:dropped_clients}
\end{figure}

\begin{figure}[H]
\centering
{\includegraphics[width=.80\textwidth]{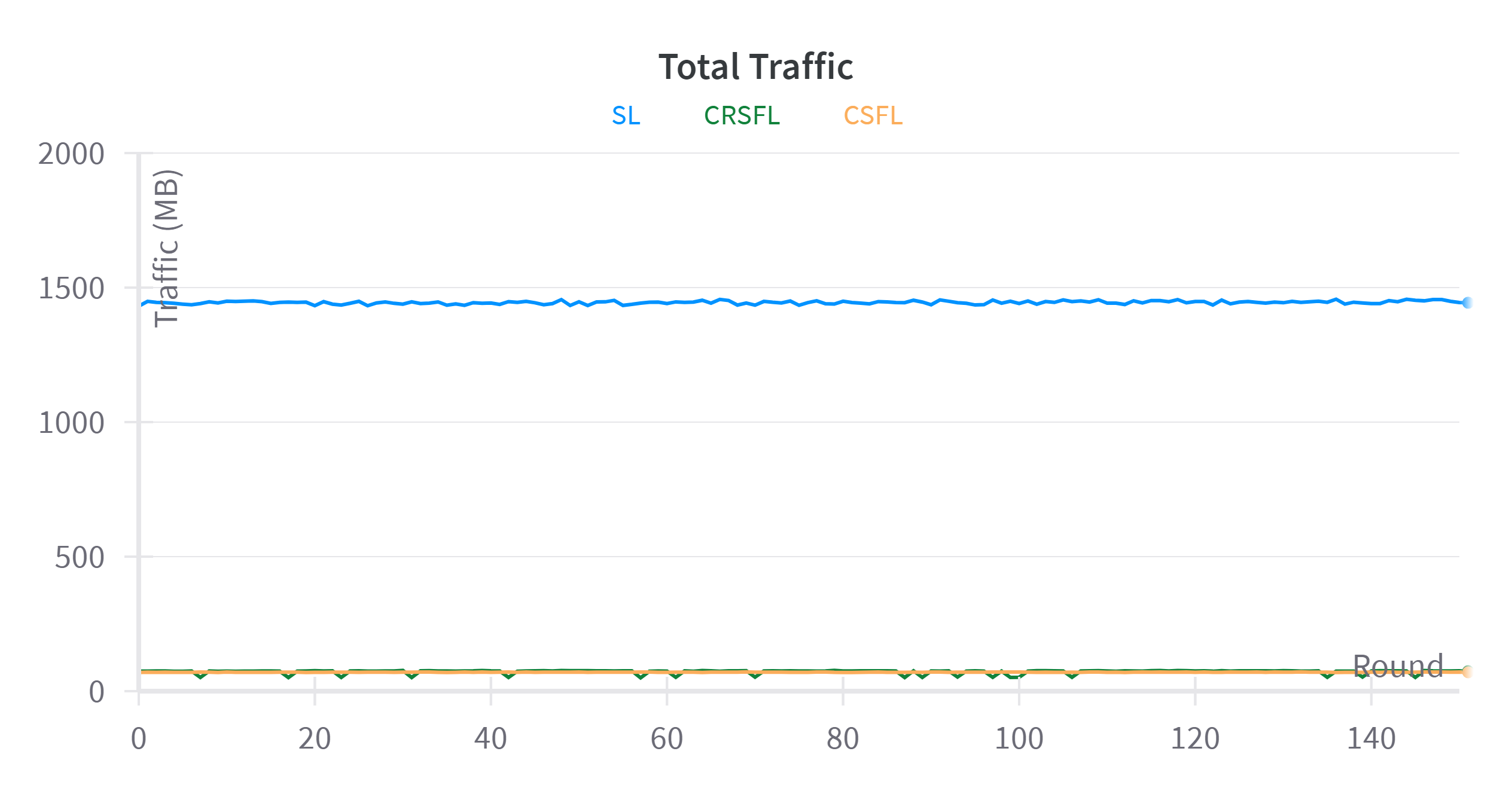}}\hfill
\caption{Clients Total Traffic Consumption in Megabytes (MB) per Round.}
\label{fig:traffic}
\end{figure}

\vspace{12pt}
\subsubsection{Client Dropout}
Figure \ref{fig:dropped_clients} discusses the number of clients dropping due to lacking resources for completing the training process; the count of the clients represents the total sum of the devices unable to complete the training process. The figure shows that the CSFL, which relies on random client selection, had the highest number of dropping clients, with an average of 8.5 clients dropping out per round. This is because the poor selection of clients who lacked the necessary resources resulted in their inability to complete the model training with the server. Conversely, with its sequential training process and fewer client selections in each round, the SL experiment had fewer clients, with an average of 4 clients dropping out per round. Finally, CRSFL has significantly fewer dropped clients than SL and CSFL due to our approach's careful selection of clients with adequate resources to complete their training. An average of 0.4 clients drop out per round. However, this small percentage of clients still drop out due to unexpected changes in their resources during model training.

\vspace{12pt}
\subsubsection{Traffic Analysis}
Figure \ref{fig:traffic} displays the total traffic of clients in megabytes (MB) during the exchange of gradients/weights with the servers, measuring communication overhead. Both CRSFL and CSFl exhibit similar traffic overhead (73 MB) since the clients share only their smashed data part of the model, and the global server-side model updates only once all clients have finished their parallel training of the model. However, SL has a much higher traffic consumption of ~1500 MB due to vanilla split learning's sequential order, where each client trains its weight and then transmits its model-side weights to the next client, significantly impacting the information exchanged between clients and servers.

\begin{figure}[H]
\centering
{\includegraphics[width=.80\textwidth]{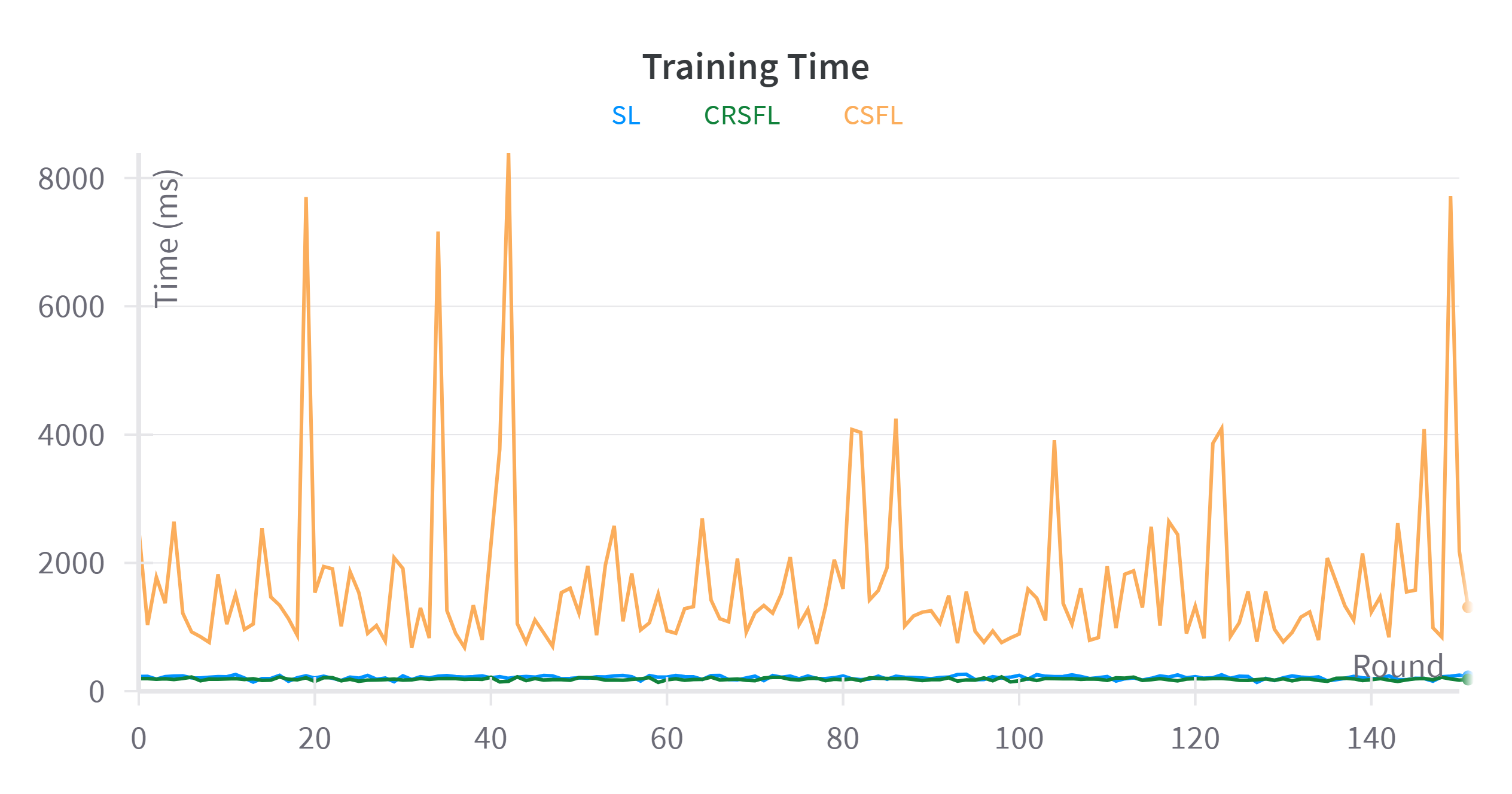}}\hfill
\caption{Clients Total Training Time in Milliseconds (ms) per Round.}
\label{fig:training_time}
\end{figure}

\begin{figure}[H]
\centering
{\includegraphics[width=.80\textwidth]{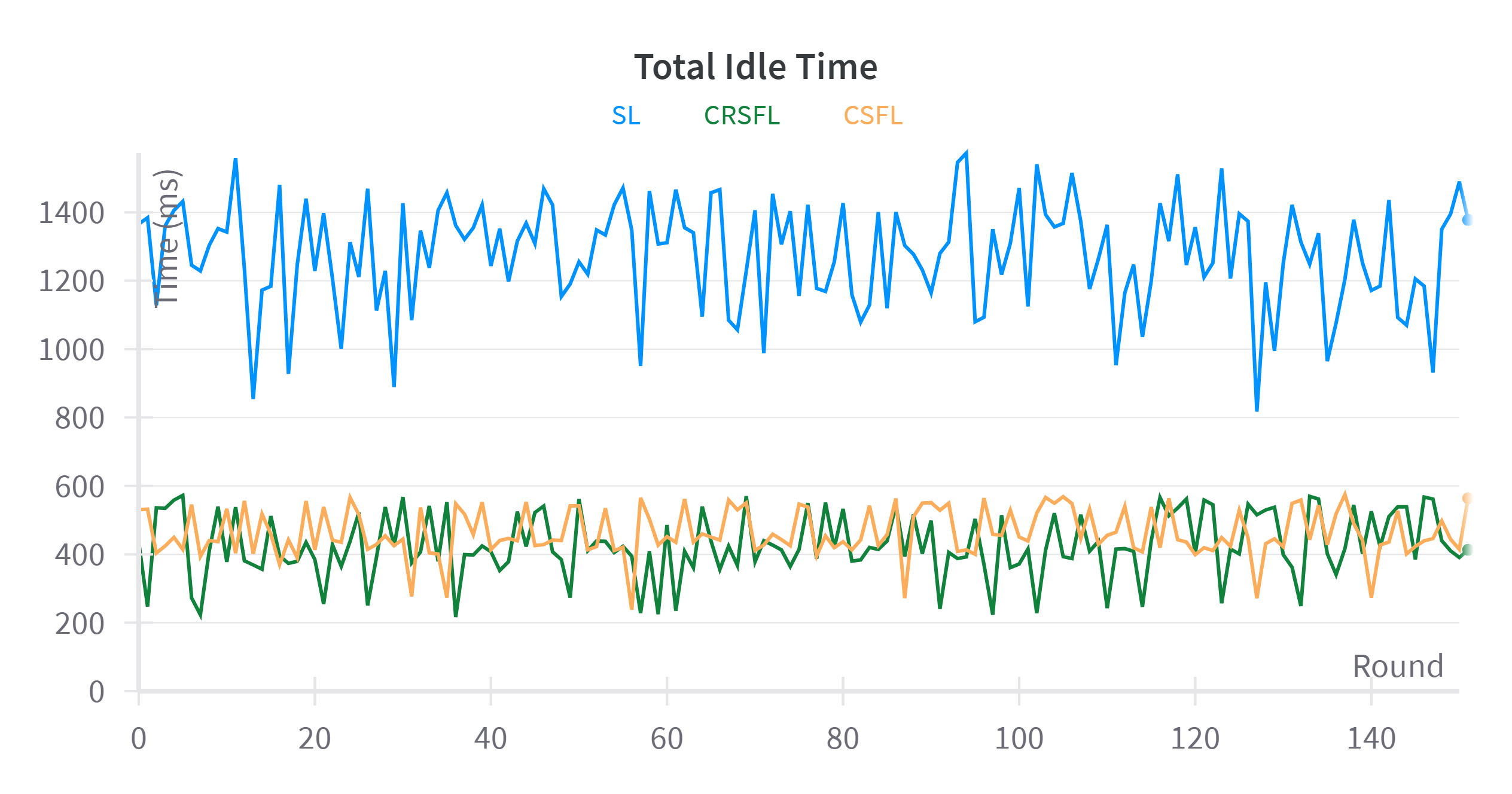}}\hfill
\caption{Clients Total Idle Time in Milliseconds (ms) per Round.}
\label{fig:idle_time}
\end{figure}
\vspace{12pt}
\subsubsection{Model Training Time}
In the graphs presented, Figure \ref{fig:training_time} displays the total training time for the selected clients per round. In contrast, Figure \ref{fig:idle_time} illustrates the total idle time per round between the chosen clients during the model training. The data shows that CSFL and CRSFL exhibit similar training times, with CSFL averaging 300 ms and CRSFL averaging 200 ms. On average, CRSFL has 100ms less idle time since both clients simultaneously train the same section of the model. CRSFL clusters the devices with similar capabilities and utilizes a GA approach that selects clients to minimize the total idle time between them during the training process. On the other hand, the idle time in SL is considerably higher, with an average of 1300ms, since the clients have to wait for each other to finish the training process.

\vspace{12pt}
Upon analysis of the data presented in the plots, our proposed approach achieves high accuracy in model convergence. It assures reduced training time, idle time for clients, and a low dropout rate. These outcomes indicate the approach's efficacy in minimizing the idle time for clients to receive system responses while maintaining high accuracy.

\section{Conclusion}
This paper presents a new technique for continuous authentication for mobile and IoT devices to verify user identity of the user continuously in the background. We proposed a Cluster-Based Resource-aware Split Federated Learning framework tailored for IoT-constrained devices. By prioritizing users' privacy through on-device data training and addressing critical challenges such as resource utilization optimization, client dropout mitigation, and efficient client selection, CRSFL offers a complete solution for split-federated learning in IoT environments. By introducing novel techniques such as filtering methodologies, clustering based on resource capacities, machine learning-based resource prediction, and heuristic client selection algorithms, we have demonstrated improvements in reducing client dropping rates, minimizing training idle time, and enhancing the efficiency of federated learning processes. Our extensive experiments on real-life authentication dataset further validate the effectiveness of CRSFL, showcasing its superiority over existing methods for continuous authentication. 

Future research aims to improve accuracy by exploring advanced model architectures like transformer networks. We strive to study the overhead burden on the client and the server while minimizing communication latency. Additional privacy-preserving techniques, such as differential privacy or secure multi-party computation, may also be investigated.

\section*{Acknowledgments}
This work was partly supported by funding from the Innovation for Defence Excellence and Security (IDEaS) program from the Department of National Defence (DND).



\bibliographystyle{elsarticle-num} 
\bibliography{references}

\end{document}